\documentclass[lettersize,journal]{IEEEtran}
\usepackage{amsmath,amsfonts}
\usepackage{algorithmic}
\usepackage{array}
\usepackage{textcomp}
\usepackage{stfloats}
\usepackage{url}
\usepackage{verbatim}
\usepackage{graphicx}
\usepackage{subcaption}

\usepackage{caption}
\usepackage{booktabs}
\usepackage{multirow}

\usepackage{array}
\usepackage{makecell}
\usepackage[compress]{cite}

\usepackage{soul} 

\newcommand{\ie}{{\em i.e.}}
\newcommand{\eg}{{\em e.g.}}

\newcommand{\etc}{{\em etc.}}

\usepackage[table]{xcolor}

\usepackage{hyperref}
\usepackage{xcolor}

\hypersetup{
    colorlinks=true, 
    linkcolor=blue,  
    citecolor=blue   
}

\def\BibTeX{{\rm B\kern-.05em{\sc i\kern-.025em b}\kern-.08em
    T\kern-.1667em\lower.7ex\hbox{E}\kern-.125emX}}
\usepackage{balance}
\begin{document}

\title{AppleVLM: End-to-end Autonomous Driving with \textbf{A}dvanced \textbf{P}erception and \textbf{Pl}anning-\textbf{E}nhanced \textbf{V}ision-\textbf{L}anguage \textbf{M}odels}

\author{Yuxuan Han$^{1}$, Kunyuan Wu$^{1}$, Qianyi Shao$^{1}$, Renxiang Xiao$^{1}$, Zilu Wang$^{1}$, Cansen Jiang$^{3}$, Yi Xiao$^{1*}$, \newline Liang Hu$^{1,2*}$ and Yunjiang Lou$^{1}$
\thanks{* Corresponding authors (Email: xiaoyi@hit.edu.cn; l.hu@hit.edu.cn).}
\thanks{$^{1}$ Y. Han, K. Wu, Q. Shao, R. Xiao, Z. Wang, Y. Xiao, L. Hu  and Y. Lou are with Shenzhen Key Lab for Advanced Motion Control and Modern Automation Equipments, Guangdong Provincial Key Laboratory of Intelligent Morphing Mechanisms and Adaptive Robotics, School of Intelligence Science and Engineering, Harbin Institute of Technology, Shenzhen, China.}
\thanks{$^{2}$ L. Hu is also with National Key Laboratory of Smart Farm Technologies and Systems, China.}
\thanks{$^{3}$ C. Jiang is with Autonomous Driving Center, Shanghai Utopilot Technology Co.Ltd., China.}

}

\maketitle

\begin{abstract}
End-to-end autonomous driving has emerged as a promising paradigm integrating perception, decision-making, and control within a unified learning framework. Recently, Vision-Language Models (VLMs) have gained significant attention for their potential to enhance the robustness and generalization of end-to-end driving models in diverse and unseen scenarios. However, existing VLM-based approaches still face challenges, including suboptimal lane perception, language understanding biases, and difficulties in handling corner cases. To address these issues, we propose AppleVLM, an advanced perception and planning-enhanced VLM model for robust end-to-end driving. AppleVLM introduces a novel vision encoder and a planning strategy encoder to improve perception and decision-making. Firstly, the vision encoder fuses spatial-temporal information from multi-view images across multiple timesteps using a deformable transformer mechanism, enhancing robustness to camera variations and facilitating scalable deployment across different vehicle platforms. Secondly, unlike traditional VLM-based approaches, AppleVLM introduces a dedicated planning modality that encodes explicit Bird’s-Eye-View spatial information, mitigating language biases in navigation instructions. 
Finally, a VLM decoder fine-tuned by a hierarchical Chain-of-Thought integrates vision, language, and planning features to output robust driving waypoints. 
We evaluate AppleVLM in closed-loop experiments on two CARLA benchmarks, achieving state-of-the-art driving performance. Furthermore, we deploy AppleVLM on an AGV platform and successfully showcase real-world end-to-end autonomous driving in complex outdoor environments.
\end{abstract}

\begin{IEEEkeywords}
End-to-End Autonomous Driving, Vision Language Model, Autonomous Vehicles, Imitation Learning
\end{IEEEkeywords}

\section{Introduction}
Autonomous vehicles provide a promising solution to improve efficiency and mobility and thus play a vital role in modern intelligent transportation systems  \cite{zhao2024autonomous,teng2023motion}. Autonomous driving systems are typically classified into two paradigms: the end-to-end and the modular pipelines. The end-to-end paradigm takes environmental observation as input and directly outputs the control signals \cite{codevilla2018end} or driving waypoints \cite{hu2023planning}, streamlining the training process and reducing data labeling costs. As a result, this paradigm has witnessed a rapid advancement in recent years \cite{chib2023recent}. 

Early end-to-end approaches were vision-based, aiming to learn driving behaviors directly from raw visual observation. Two widely used learning methods in this paradigm are: \textit{Reinforcement Learning (RL)} \cite{zhang2021end} and \textit{Imitation Learning (IL)} \cite{lu2023imitation}. RL-based approaches learn driving policies through trial and error using handcrafted reward functions, whereas IL leverages expert demonstrations to map observations to driving actions without explicitly defining a reward function. IL does not face the issues of sparse rewards and does not require manually defining an explicit reward function (which can be extremely complicated in certain scenarios), and hence has shown promising performance in autonomous driving \cite{le2022survey}. 

However, relying solely on visual input is insufficient for real-world driving. For instance, when approaching an intersection, navigation instruction is essential to indicate the correct direction; similarly, in emergency scenarios such as pedestrians crossing, contextual warnings can improve safety \cite{pan2024vlp}. To address such limitations, Vision Language Models (VLMs) have been introduced in autonomous driving models~\cite{zhou2024vision}, integrating the inputs of both vision and language modalities. By leveraging the power of foundation models such as CLIP \cite{radford2021learning} and Large Language Models (LLMs) \cite{touvron2023llama}, the VLMs enhance understanding and reasoning in complex driving scenarios \cite{liu2023improvedllava}. Typically, a VLM-based driving framework consists of three modules: 1) an input encoder that tokenizes the vision and natural language inputs as a sequence of multi-modal representation; 2) a language model that processes these tokens to generate driving decisions in text form; 3) an information decoder that converts these decisions into driving waypoints and control signals for driving~\cite{cui2024survey}. In contrast to Vision-Language-Action (VLA) models~\cite{zhou2025autovla1,zhou2025opendrivevla} that directly regress actions (\textit{e.g.}, control signals) from inputs, we adopt VLMs along with an extra controller to provide explicit reasoning and improve interpretability~\cite{jiang2025survey}.

LMDrive \cite{shao2024lmdrive} pioneered the use of LLMs for closed-loop autonomous driving. Though the existing end-to-end autonomous driving models show strong capabilities to handle complex instructions and challenging driving scenarios in the CARLA simulator, they still struggle in real-world applications~\cite{chen2024driving, cao2024maplm}. Specifically, end-to-end autonomous driving models are sensitive to sensor configuration such that even minor variations in sensor placements and resolutions can significantly impact perception, limiting the scalability across different vehicles. Furthermore, the natural language, as an abstract expression, lacks spatial precision and might cause understanding ambiguity. For instance, an instruction like “Go straight at the intersection” may not specify which lane to take when merging onto a multi-lane road, causing unintended lane oscillations. Moreover, current VLM-based driving models are pre-trained on standard driving datasets, making them ineffective in handling rare yet critical cases such as a sudden pedestrian crossing.

To overcome these challenges, we propose a novel model called AppleVLM, short for \textbf{A}dvanced \textbf{P}erception and \textbf{PL}anning-\textbf{E}nhanced
Vision-Language Models for end-to-end autonomous driving. AppleVLM follows the IL method, trained from a set of expert driving data for closed-loop driving. Our contributions can be summarized as follows:

\begin{enumerate}
     \item{We introduce a novel end-to-end driving framework that is more robust to sensor configurations and resolutions, highlighting its potential as a scalable solution for real-world deployment across diverse vehicle platforms;}
     \item{The proposed vision encoder employs a deformable attention mechanism to fuse RGB images and point-cloud data across both spatial (multi-view) and temporal (multi-timestep) dimensions, which significantly improves lane perception accuracy and enhances robust waypoint prediction;}
	\item{We propose a planning strategy encoder to explicitly represent driving scenarios. With the planning features, the model can significantly mitigate language-induced perception biases and improve interpretability;}
	\item{The real-world corner-case datasets are used to fine-tune the VLM backbone with Chain-of-Thought (CoT). It enables our model to involve more out-of-distribution data, thus achieving better generalization performance in both simulation and the real world.}
\end{enumerate}   

\section{Related Work}
\subsection{End-to-End Imitation Learning}
End-to-end autonomous driving has gained significant attention due to its simplicity in training and the benefit of not requiring manual data labeling \cite{lu2023imitation,jaeger2023hidden}. Current end-to-end driving models apply either RL or IL methods \cite{al2024end,zare2024survey,chen2019deep}. Designing effective reward functions to address diverse scenarios in open-world settings remains a challenge in RL, and its initial training and testing are quite dangerous. In comparison, IL has been more widely studied as an alternative.

Existing end-to-end IL models in autonomous driving have two types of output: action and waypoint. The action-based end-to-end driving models \cite{codevilla2018end,9165167,10341506} directly map observation to vehicle control action, offering a straightforward solution without requiring controller design. On the other hand, the waypoint-based end-to-end driving models \cite{Chitta2023PAMI, shao2023safety, hu2023planning, jaeger2023hidden} output waypoints from observation, and then a controller is designed and tuned to transform trajectory waypoints to vehicle control, which provides a clearer understanding of decision-making and better diagnostics.
To take both advantages, TCP \cite{wu2022trajectoryguided} integrates trajectory planning and direct control by two branches, with one predicting future waypoints while the other one reasoning multi-step actions. 
Recently proposed Transfuser++ \cite{jaeger2023hidden} discusses these two types of end-to-end models and achieves more robust results in the waypoint-based ones. Therefore, in this work, we focus on the waypoint-based end-to-end models.

Early end-to-end driving models rely on uni-modal vision input. Transfuser \cite{Chitta2023PAMI} pioneers the deep fusion of image and point-cloud features using hybrid transformer-CNN architectures, achieving remarkable waypoints prediction results on the Longest6 \cite{Jia2024NeurIPS} benchmark. InterFuser \cite{shao2023safety} further improves accuracy by using a single transformer for vision encoding and waypoint decoding. UniAD \cite{hu2023planning} is the first work of full-stack driving architecture that generates waypoints for vehicle navigation in the real world.
These vision-based uni-modal models demonstrate the feasibility of end-to-end driving, however, they only focus on feature extraction in images but lack the direct understanding of semantic information, which makes it difficult to understand abstract concepts in complex scenes. 
To bridge this gap, multi-modal vision-language models have emerged as promising paradigms, including VLA and VLM~\cite{jiang2025survey}. VLA models~\cite{zhou2025autovla1, zhou2025opendrivevla} that directly map vision-language inputs to actions have gained traction for their training simplicity. In contrast, the VLM-based approaches provide explainable decision-making through natural language and planning strategy, as summarized in~\cite{jiang2025survey}.

\subsection{VLMs in Driving}
In recent years, VLMs have attracted widespread attention due to their ability to integrate multi-modal information ({\ie} vision and language) for better understanding and reasoning in autonomous driving~\cite{ma2025position,cui2024receive, cui2024board}.
DriveGPT-4~\cite{xu2024drivegpt4} introduces a multi-modal VLM that takes a sequence of image frames and human questions as input, and outputs control signals and answers. However, it lacks the input of navigation instructions, thus cannot drive following planned routes. Moreover, it was evaluated on open-loop datasets, which might cause the covariate shift problem that occurs gradually over time or during real-world deployment.
DriveLM~\cite{sima2025drivelm} builds upon nuScenes and CARLA, and proposes a VLM-based approach for jointly performing graph visual question-and-answer and end-to-end driving. 
DriveVLM~\cite{tian2024drivevlm} takes images as input and outputs scene description, scene analysis, and hierarchical planning results through a CoT mechanism, achieving enhanced scenario understanding.
LMDrive~\cite{shao2024lmdrive} is the very first work to leverage LLMs and shows promising performance in closed-loop end-to-end driving. 

These works have demonstrated the promising potential of VLMs in end-to-end driving, however, several limitations remain to be addressed for performance improvement. Firstly, existing VLMs for autonomous driving exhibit high sensitivity to sensor configurations during training and testing. In other words, even minor variations in sensor positioning and orientation can significantly impact driving performance. 
Secondly, VLMs exhibit inherent language biases, where ambiguous or imprecise language expressions may lead to misinterpretations of driving scenarios. Incorporating more explicit feature representations may mitigate the issue. 
Moreover, current closed-loop VLMs do not cover corner-case driving scenarios in model training, which is more likely to lead to covariate drift in real-world driving.

\begin{figure*}[!t]
	\centering
	\includegraphics[width=\textwidth]{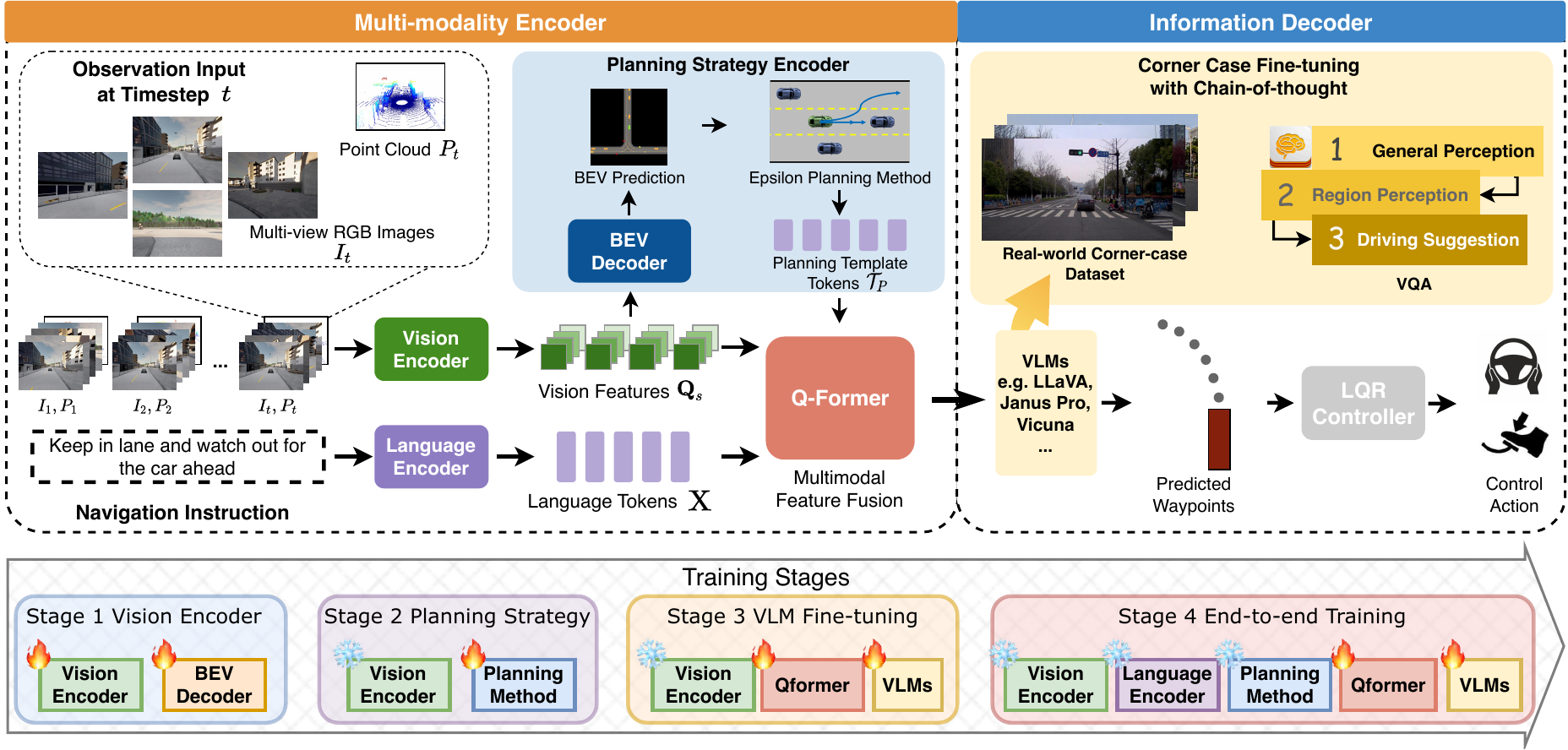}
	\caption{
    The proposed AppleVLM follows an encoder-decoder architecture: 
    The \textbf{Multi-modality Encoder} includes three types of encoders: 1) a vision encoder processes a time sequence of multi-view sensor data (RGB images and point-cloud) and generates vision features; 2) a language encoder encodes the navigation instructions to language tokens; 3) a planning strategy encoder takes vision features as input and outputs the planning template tokens. These features from three modalities are fused by a module based on the Q-Former architecture. The \textbf{Information Decoder} adopts a VLM backbone (such as LLaVA or Janus Pro) to process the multi-modal features. This VLM is pre-trained with corner-case data following a CoT mechanism with three tasks: general perception, region perception, and driving suggestion to predict a sequence of driving waypoints. During the training process of end-to-end driving, the fine-tuned VLM is frozen, and an LQR controller is adopted to transform waypoints to control actions for actual driving, {\ie} steering angle, throttle and brake. The training process consists of four stages: \textbf{stage 1} is pre-training the vision encoder for BEV prediction; \textbf{stage 2} is learning the planning strategy with features from the frozen vision encoder; \textbf{stage 3} is fine-tuning the VLM with Q-Former on corner-case data with CoT mechanism; and in \textbf{stage 4}, the end-to-end training leveraging features from all frozen encoders, and the trainable Q-Former and VLM.}
	\label{fig_overall}
\end{figure*}

\section{Methodology} \label{methodology}
\subsection{Problem Setting}\label{sec:ps}
To train the AppleVLM, we follow the IL method, where the idea is to train an agent that drives with a policy $\pi$ that imitates the policy of an expert driver $\pi^*$. The policy $\pi$ is trained in a supervised manner following the behavior cloning \cite{zhao2024autonomous} approach of IL. In this work, as we focus on waypoint-based end-to-end driving, the policy is a mapping from observation to waypoints, which are then fed to a separate controller and output driving control actions.

Before training an agent, we allocate an expert driver in the environment to generate a dataset $\mathcal{D}=\{(\mathbf{O}_l, \mathbf{W}_l)\}^L_{l=1}$ of size $L$, which consists of high-dimensional observation of the environment $\mathbf{O}$, and the corresponding expert driving trajectories represented by a set of 2D waypoints in Bird's-Eye-View (BEV) space $\mathbf{W}=\{(x_t,y_t)\}^T_{t=1}$ where $T$ is the number of predicted waypoints at each time step. This dataset is then used to train a parameterized policy $\pi_{\theta}(\cdot)$, which approximates the expert policy. The training objective is:

\begin{equation} \label{eq:target}
  \arg\min_{\theta} \mathbb{E}_{(\mathbf{O}_{l}, \mathbf{W}_{l})\sim \mathcal{D}} \left[ \mathcal{L}(\pi_\theta(\mathbf{O}_{l}), \mathbf{W}_{l}) \right].
\end{equation}

Overall, the training process of AppleVLM consists of four stages: 1) \textit{Vision Encoder Pre-training with BEVs}: we use BEV prediction as a pre-training task for the vision encoder, aiming to encode spatial information for better vision feature extraction; 2) \textit{Planning Strategy Encoder Training}: the output features of the vision encoder is used as input to train the planning strategy encoder, where a set of spatial-temporal corridors are generated; 3) \textit{VLM Fine-tuning with Corner Cases}: we use a real-world corner-case dataset to fine-tune the VLM baseline to alleviate the long-tail problem of training data; 4) \textit{End-to-end Training of AppleVLM}: the pre-trained vision encoder and planning strategy encoder are frozen, while the rest part of the model is end-to-end trained for the driving task. The corresponding losses are described in \ref{sec:td}.

\subsection{Architecture}\label{sec:archi}
As shown in Fig.~\ref{fig_overall}, the proposed AppleVLM comprises a \textit{Multi-modality Encoder} and an \textit{Information Decoder}. 
We will introduce these two modules in \ref{sec:mme} and \ref{sec:id}, respectively,  and finally describe the training details
in \ref{sec:td}.
\subsubsection{Multi-modality Encoder}\label{sec:mme}
AppleVLM consists of encoders that incorporate features from three modalities: vision, language, and planning strategy.

\begin{figure*}[!t]
	\centering
	\includegraphics[width=\textwidth]{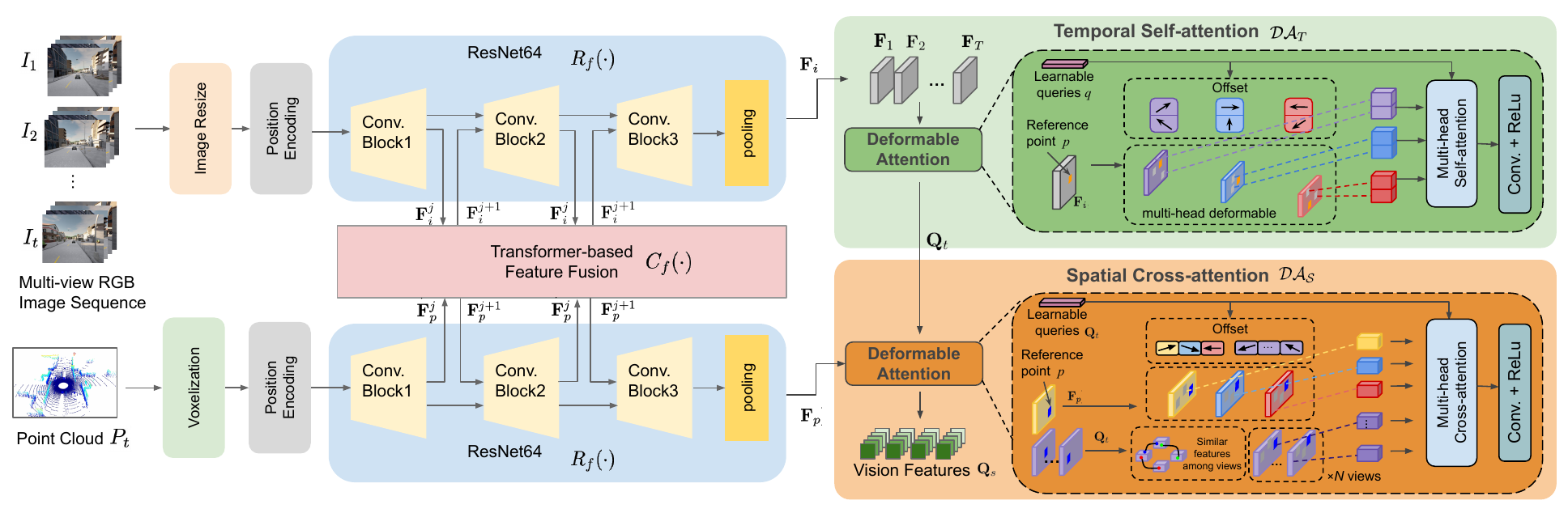}
	\caption{Details of the vision encoder. The features of images and the point cloud are fused by the self-attention mechanism at several convolution blocks in the ResNet64 backbone. Furthermore, a deformable self-attention mechanism is applied to the image feature sequence over $T$ frames, and a deformable cross-attention mechanism is adopted to associate features from the modalities of images and the point cloud.
    }
	\label{fig_vision_backbone}
\end{figure*}

\noindent \textbf{Vision Encoder:}
Fig.~\ref{fig_vision_backbone} shows the details of the vision encoder. The perception input $\mathbf{O}$ consists of a sequence of multi-view RGB images $\mathbf{I} =\{ \{ I_{t,1}, I_{t,2}, \dots, I_{t,N} \}_{t=1}^T \}$ ($T$ and $N$ are numbers of time-step and view), and the point cloud data  $\mathbf{P}_ = \{ P_1, P_2, \ldots, P_t \}_{t=1}^T$.
To align the dimension of perception input data, we apply bilinear interpolation on $\mathbf{I}$ to normalize the image size, and voxelization on $\mathbf{P}$, which is further processed to pseudo image $\mathbf{I}_{p}=\phi(\mathbf{P})$. We use two RegNet64 backbones $R_{f}(\cdot)$ to extract features $\mathbf{F}_{i}$ and $\mathbf{F}_{p}$ from $\mathbf{I}$ and $\mathbf{I}_{p}$ respectively, described as below:
\begin{equation}
\begin{aligned}
    \mathbf{F}_{i} = R_{f}(\mathbf{I}), \; \mathbf{F}_p = R_{f}(\phi(\mathbf{P})), \\
    \mathbf{F}^{j+1}=[\mathbf{F}_{i}^{j+1}, \mathbf{F}_p^{j+1}] = C_f(\mathbf{F}_{i}^j, \mathbf{F}_p^j).
\end{aligned}
    \label{feature_cross}
\end{equation}

For each convolutional block in $R_{f}(\cdot)$, we take the feature output at current block $j$, $\mathbf{F}_{i}^{j}$ and $\mathbf{F}_p^{j}$, and consecutively obtain $\mathbf{F}_{i}^{j+1}$ and $\mathbf{F}_p^{j+1}$ by the feature fusion function $C_f(\cdot)$ in the Transformer with shared weights. Through the cross-attention mechanism, we fuse the features from different modalities of vision input.

Once the fused feature $\mathbf{F}=[\mathbf{F}_i, \mathbf{F}_p]$ 
is obtained via  \eqref{feature_cross}, we further process it through a temporal self-attention block
and a spatial cross-attention block 
built upon the Deformable Attention ($\mathcal{DA}$) \cite{zhu2020deformable}, which is formally described as below:
\begin{equation*}
    \mathbf{Q} = \mathcal{DA}(q, p, \mathbf{F})=\sum_{m=1}^{M}\sum_{k=1}^{K} \mathcal{A}_{mk} \cdot \mathcal{W}_{m}(q) \cdot \mathbf{F}\left(p+\Delta p_{w,h}\right)
    \label{temporal_deformable}
\end{equation*}
where $q$ is the query, $p$ is the reference point, and $\Delta p_{w,h}$ represents the predicted offsets regarding to $p$, 
$M$ and $K$ indicate the numbers of the attention heads and the sampled keys in each head, respectively.
$\mathcal{W}_{m}$ represents the learnable weights in head $m$
, and $\mathcal{A}_{mk}$ is the predicted attention weight. This enhanced attention mechanism dynamically adjusts the region of interest by introducing deformable offsets and thus can flexibly capture important features across multiple heads.

Based on the idea of $\mathcal{DA}$, for $\mathbf{F}_i$ that contains features of the $T$ timesteps, we apply a temporal self-attention block $\mathcal{DA}_{T}$ to associate features among the sequence and obtain $\mathbf{Q}_t$ as below:
\begin{equation*}
    \mathbf{Q}_t = \mathcal{DA}_{T}(q, p_f, [\mathbf{F}_i^{1},\dots,\mathbf{F}_i^{T}]\}) =\sum_{t=1}^{T}  \mathcal{DA}(q, p_f, [\mathbf{F}_i^{t}]),
    \label{temporal_feature}
\end{equation*}
then the temporal fused feature of all views $\mathbf{Q}_t$, along with the feature output of the point cloud branch $\mathbf{F}_p$ are fed into a spatial cross-attention block $\mathcal{DA_S}$:
    \begin{align*}
                    \mathbf{Q}_{s} &= \mathcal{DA}_{S}(\mathbf{Q}_t, p_f, (\mathbf{Q}_t,\mathbf{F}_p)) \\
                    &= (\frac{1}{|\mathcal{V}_{\text{hit}}|}\sum_{m\in\mathcal{V}_{\text{hit}}}\mathcal{DA}(\mathbf{Q}_t, p_f, \mathbf{Q}_t \in m))+\mathcal{DA}(\mathbf{Q}_t, p_f, \mathbf{F}_p)
    \end{align*}
    \label{spatial_feature}
In this block, we take the $\mathbf{Q}_t$ as the query, and $m$ views with similar features in $\mathbf{Q}_t$, {\ie} $\mathcal{V}_{\text{hit}}$, as the key, to apply the cross-attention feature fusion.
Next, the same $\mathcal{DA}$ is applied while using the point cloud feature output $\mathbf{F}_p$ as the key.
Finally, these features are concatenated as the final feature output of the vision encoder, $\mathbf{Q}_s$.

Notably, our proposed vision encoder can adapt variations in sensor configurations ({\ie} position, orientation) due to the deformable attention mechanism, which adaptively adjusts the sampling position to capture changes in the scene. This bypasses cumbersome camera calibration and increases model generalization during real-world deployment.

\noindent \textbf{Language Encoder:}
The navigation instruction is built upon the LMDrive, which consists of three specific scenarios ({\eg} follow, turn, and others) \cite{shao2024lmdrive}. 
To process the input data of the language encoder, we follow the conventional natural language processing procedure, splitting the navigation instruction into tokens, such as words and sub-words, and removing the punctuation. Then, these tokens are mapped to a vector of input sequence $\mathbf{{X}} = (x_1, x_2, \dots, x_n)$ by Word2Vec \cite{mikolov2013efficient}
Finally, special tokens ({\eg} [CLS], [Distance]) are added to the sequence $\mathbf{X}$ as specific features for autonomous driving.

\noindent \textbf{Planning Strategy Encoder:}
Different from most existing VLM-based autonomous driving models, AppleVLM incorporates a novel planning strategy encoder that leverages the explicit representation of the environmental information, {\ie}, lane perception and the states of surrounding objects, to reduce the bias caused by the implicit representation of navigation instructions in natural languages.
Specifically, we decode the output feature of the vision encoder $\mathbf{Q}_s$ with a BEV decoder consisting of several convolutional layers and multi-layer perceptron (MLP) to output BEV semantic segmentation images $\mathbf{B}\in [0,1]^{H_g \times W_g \times C_g}$, consisting of $C_g$ grayscale images of size $H_g \times W_g$. In our case, we consider $Cg=3$, including semantic grayscale images of the drivable area $\mathcal{X}$, the lane boundary $\mathcal{E}$, and the state of the surrounding objects $\mathcal{S}$. 

\begin{figure}[!t]
    \centering
        \includegraphics[width=0.48\textwidth]{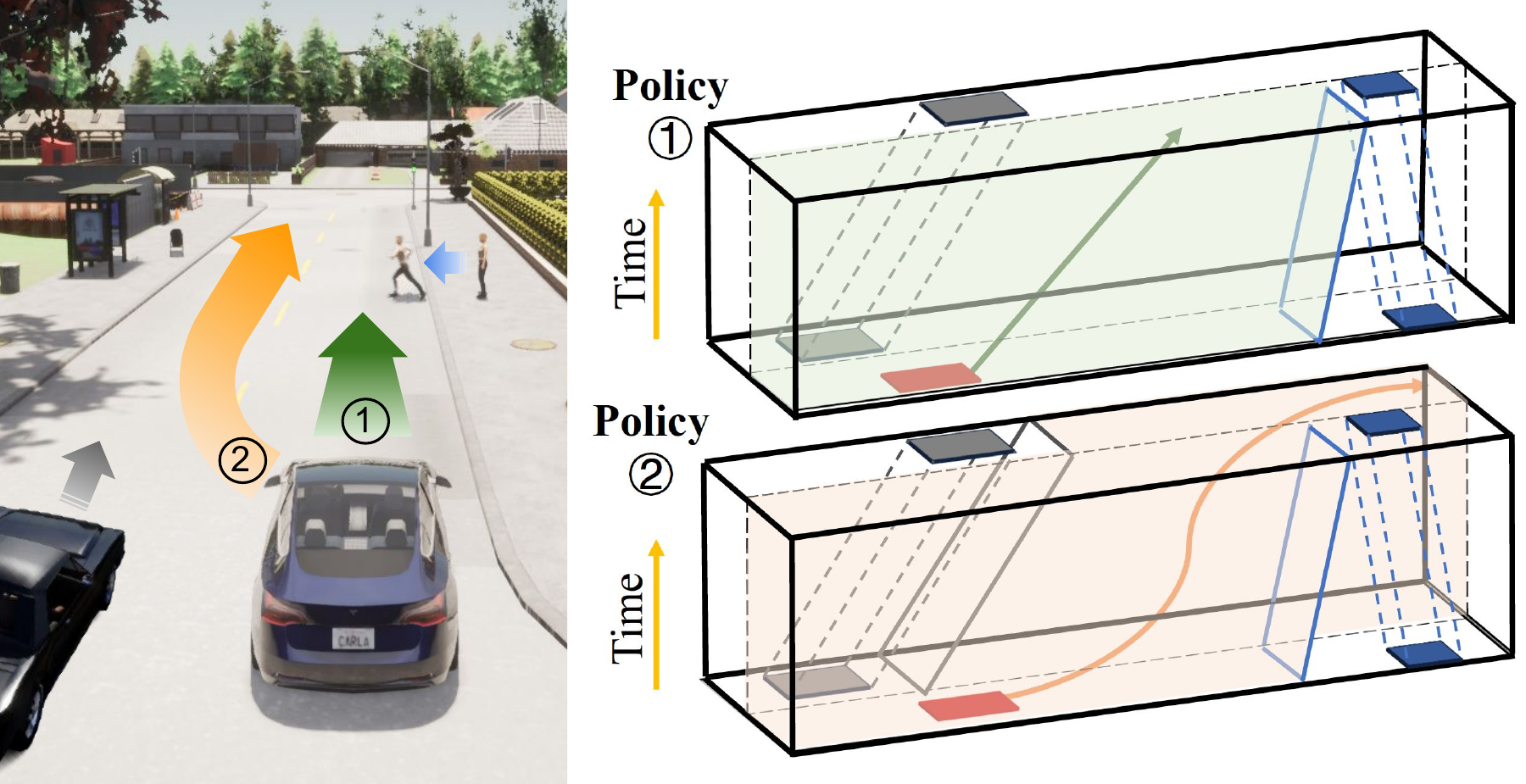}
        \caption{An example of a driving scenario is shown on the left. By applying the Epsilon \cite{ding2021epsilon} planning method, the corridors that represent the possible driving space along time are generated on the right. Corridors of the other vehicle and pedestrian are illustrated in gray and blue respectively. The red rectangle indicates the initial location of the ego vehicle. By integrating the top $N$ policies (the orange and green lines), we conduct a constraint corridor for the ego vehicle.}

    \label{fig:ssc1}
\end{figure}

Based on this environmental information, we adopt the Epsilon \cite{ding2021epsilon} planning method $\mathcal{P}_e(\cdot)$ to generate driving trajectories and candidate driving corridors with high scores of success and safety in the scene, as depicted in Fig.~\ref{fig:ssc1}. 
Specifically, the top $N$ policies $\mathbb{T}^p$ generated by $\mathcal{P}_e(\cdot)$ are selected, that is  
\begin{equation*}
        \mathbb{T}^p= \arg\max_{\substack{x_{t_H} \in \{x_{t_H}^1, x_{t_H}^2, \dots, x_{t_H}^i\}}}^{N} (\mathcal{P}_e(\mathcal{X},\mathcal{E},\mathcal{S}))
    \label{top3_equation}
\end{equation*}
This representation explicitly offers temporal and spatial geometry information of the driving scene, thereby mitigating language biases induced by abstract navigation instruction.

With $\mathbb{T}^p$, we transform them into a three-dimensional tensor $\mathbb{P}^p = [T,X,Y]$, where the temporal dimension $T$ and the spatial occupancy grid $[X,Y]$ of corridors at time $T$ are encapsulated.
We concatenate different policies tensor as a whole in $[\mathbb{P}^p]$. Then, we apply a Spatial-Temporal Transformer Tokenizer \cite{li2024multi} $\textbf{Token}(\cdot)$ to encode $[\mathbb{P}^p]$ to planning template tokens $\mathbf{\mathcal{T}}_{P}$, that is $\mathbf{\mathcal{T}}_{P} = \textbf{Token}([\mathbb{P}^p]) \; \textrm{where} \; p \in N
    \label{equal_tokenizer}$.

\begin{figure}[!t]
	\centering
	\includegraphics[width=8cm]{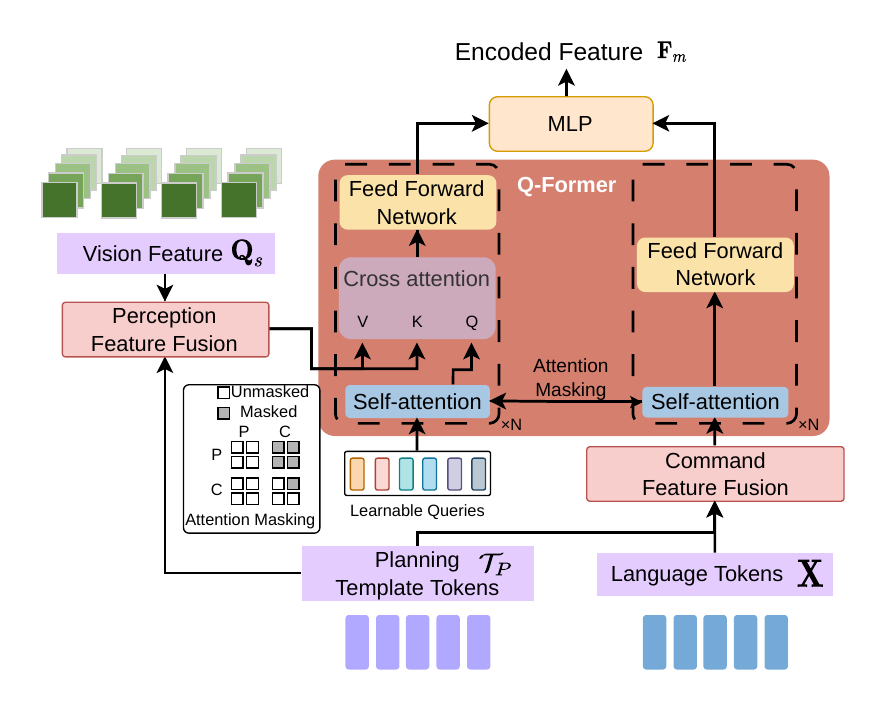}
	\caption{
    The planning template tokens $\mathbf{\mathcal{T}}_{P}$ encoded from explicit representation ({\ie} the spatial-temporal corridor) of the driving scene are fused into the vision and language features by a Q-former-based architecture.
    }
	\label{fig_Qformer}
\end{figure}

\noindent \textbf{Multi-modality Feature Fusion:}
As illustrated in Fig.~\ref{fig_Qformer}, we integrate vision features, language tokens, and planning template tokens by a Querying Transformer (Q-Former) \cite{li2023blip} architecture. 
We first apply cross-attention $\mathcal{A}_{cross}(\cdot)$ to obtain two fused features: \textit{perception} feature $\mathbf{F}_{perception}$ that are fused by vision features and planning template tokens, and \textit{command} feature $\mathbf{F}_{command}$ that are fused by language and planning template tokens as below:
        \begin{align*}
                    \mathbf{F}_{perception} &= \mathcal{A}_{cross}(\mathbf{Q}_s, \mathbf{\mathcal{T}}_{P})  \\
                    \mathbf{F}_{command} &= \mathcal{A}_{cross}(\mathbf{X}, \mathbf{\mathcal{T}}_{P})
        \end{align*}
Then $\mathbf{F}_{perception}$ and $\mathbf{F}_{command}$ are fed into the Q-Former with the learnable queries. The idea is to extract meaningful and most relevant features from $\mathbf{F}_{perception}$ and convert them into a format that can be effectively fused with $\mathbf{F}_{command}$, achieving the alignment of the features from different modalities. Finally, the outputs of the Q-former are concatenated and passed to an MLP module to obtain encoded feature $\mathbf{F}_m$.

\subsubsection{Information Decoder}\label{sec:id}
Before the end-to-end training of AppleVLM, the VLM backbone is pre-trained
through a CoT mechanism with three tasks: general perception, region perception, and driving suggestion. When training the end-to-end AppleVLM, the fine-tuned VLM is frozen, and the encoded feature $\mathbf{F}_m$ is fed into the VLM. Then, the VLM outputs a set of driving waypoints, which are further processed by a control model to produce vehicle control actions: steering angle, throttle, and brake.


\noindent \textbf{VLM pre-trained with CoT:} 
We choose the latest proposed model, Janus Pro \cite{chen2025janus}, as the VLM backbone. Considering that autonomous driving models need to be rolled out in a wide range of driving scenarios, including rare or unexpected situations, we fine-tune the VLM with driving-specific corner-case data \cite{li2024automated} that is not well-represented in standard training datasets. Inspired by \cite{li2024automated} and \cite{sima2025drivelm}, the fine-tuning contains a CoT process with three tasks: 1) \textit{General Perception}, where the task is to understand which kinds of critical entities are on the road, and the reasons why they influence the driving behaviors of the ego vehicle; 
2) \textit{Region Perception}, where the task is to describe objects within the given bounding boxes and explain why they would influence the ego vehicle's behavior; and 3) \textit{Driving Suggestion}, where the aim is to suggest a driving trajectory represented by a set of waypoints based on the results from previous tasks. 

\begin{figure*}
    \centering
    \includegraphics[width=\textwidth]{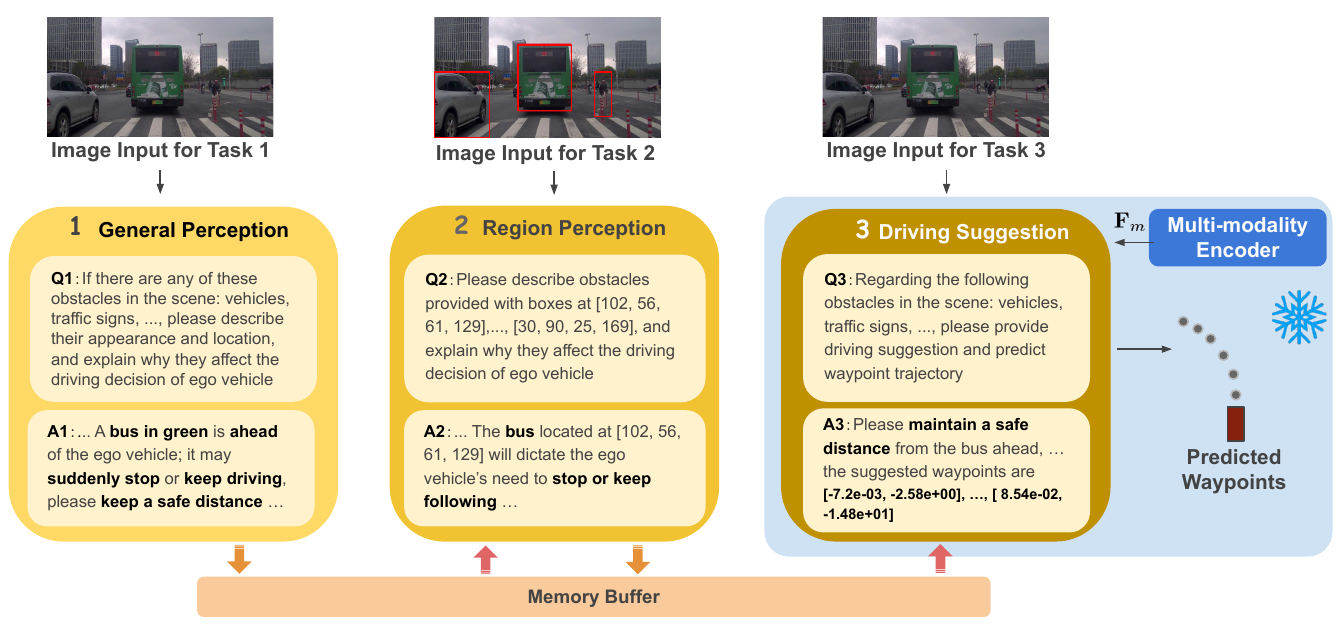}
   \caption{The flow of VLM fine-tuned with CoT. 
   The process formulates reasoning as a series of question-answer pairs in
   three tasks: General Perception, Region Perception, and Driving Suggestion. During the fine-tuning, the output of each task is saved into a memory buffer and further used for the following ones. The fine-tuned VLM is frozen during the end-to-end training and inference time of AppleVLM, which is demonstrated in the blue region.}
    \label{fig:training}
\end{figure*}

Fig.~\ref{fig:training} illustrates the CoT process with an example. During the fine-tuning, the output answer of each task, as the chain, is concatenated in the prompt for the following tasks. By fine-tuning explicit tasks, the VLM hierarchically enriches its vision-language pairs, thereby enhancing its reasoning capabilities.
Note that during the training and inference time of the end-to-end AppleVLM, we only adopt the last task of the CoT, {\ie}, inputting the encoded feature $\mathbf{F}_m$ and the question of \textit{Driving Suggestion} to the VLM to generate the suggested driving waypoints.
This significantly improves the computational efficiency of AppleVLM and allows it to run smoothly on constrained hardware platforms without requiring excessive computing resources, enabling real-time running in real-world deployment.


\noindent \textbf{Control Model:} Once the driving waypoints are predicted, we utilize a dynamic control model with the Linear Quadratic Regulator (LQR) algorithm \cite{1017552} to output the control signals for driving: steering, throttle and brake.

\subsection{Loss Function}\label{sec:td}
The whole training process of AppleVLM consists of three stages and hence different loss functions are defined at different stages:

\subsubsection{Vision Encoder Pre-training with BEVs}
To enhance the spatial understanding of driving scenes, we pre-trained our vision encoder using BEV data covering three semantics: surrounding agents, lane boundaries, and drivable areas. The loss function consists of three parts.

\noindent \textbf{Classification Loss:}
for classification of $C_s$ categories, the focal loss $\mathcal{L}_{cls}$ is defined as below: 
\begin{equation*}
    \mathcal{L}_{cls} = - \sum_{c=1}^{C_s} \alpha_c (1 - p_{c})^{\gamma_c} \log(p_{c}) 
    \label{bev_cls}
\end{equation*}
where $p_c$ represents the predicted probability of category $c$; $\alpha_c$ is the weight factor to alleviate the class imbalance; $\gamma_c$ is the parameter of modulation factor $(1-p_c)^{\gamma_c}$, which is used to control the contribution of classifying category $c$ to the loss. A larger $\gamma_c$ reduces the loss of categories with high $p_c$, forcing the model to pay more attention to the hard misclassified categories. In our case, In our case, we set $C_s=7$, including categories of vehicle, pedestrian, red light, road lane, lane marking, sidewalk, and unlabeled, and we consider two hard-to-classify categories: pedestrian and lane marking.

\noindent \textbf{Road Structure Loss:}
The cross-entropy loss is used for classifying the lane boundary and drivable area and background:
\begin{equation*}
    \mathcal{L}_{road} = -\frac{1}{N} \sum_{i=1}^{N} \sum_{c_d=1}^{C_d} y_{i,c_d} \log(p_{i,c_d})
    \label{bev_hd}
\end{equation*}
where $C_d=3$, $p_{i,c_d}$ denotes the predicted probability of grid pixel $i$ belonging to category $c_d$, and $y_{i,c_d}$ is the ground truth.
\noindent \textbf{Occupancy Loss:} $\mathcal{L}_{occ}$ is the binary focal loss applied on the BEV map for classifying background/foreground objects, defined as below: 
\begin{equation*}
    \mathcal{L}_{occ} = \alpha_t (1 - p_t)^{\gamma} \log(p_t)
\label{bev_loc}
\end{equation*}
where $p_t$ indicates the predicted probability of a pixel occupied by foreground objects, and the weight factor $\alpha_t$ and the parameter $\gamma$ of modulation factor are adjusted to force the model to focus more on the foreground pixels.



\subsubsection{Planning Strategy Encoder Training}
Following the EPSILON~\cite{ding2021epsilon} method, we train the planning strategy encoder via imitation learning and freeze it during inference. Given a state sequence $s_t$, the objective is to minimize the negative log-likelihood of the  action distribution of the policy with respect to expert demonstrations, described by:

\begin{equation}
    \mathcal{L}_{PS} = -\sum_{t=1}^{T} \log \pi_\theta(a_t^{\text{expert}} \mid s_t),
    \label{loss_ep}
\end{equation}
where $\pi_\theta$ is the policy network and $T$ is the trajectory length.

\subsubsection{VLM Fine-tuning with Corner Cases}
When fine-tuning the VLM decoder, we apply negative log-likelihood loss function $\mathcal{L}_{\text{NLL}}$ as below:
\begin{equation*}
    \mathcal{L}_{NLL} = -\sum_{t=1}^T \log p(y_t \mid y_{<t}, x)
    \label{NLL}
\end{equation*}
where $x = (x_1, x_2, \dots, x_T)$ is the input prompt tokens and $y = (y_1, y_2, \dots, y_T)$ is the target output sequence. For a sequence with the length of $T$, the goal of the model is to generate sequence $y_{<t} = (y_1, y_2, \dots, y_{T-1})$ 
that approximates to the target sequence $y$.

\subsubsection{End-to-end Training of AppleVLM}
Once we obtain the pre-trained vision encoder and the fine-tuned VLM decoder, we freeze these two modules and further train the AppleVLM for driving using an $L_1$ loss between the predicted waypoints and the ground truth waypoints. Let $\mathbf{w}_k^\star$ and $\mathbf{w}_k^{gt}$
 represent the predicted and ground truth waypoints at
timestep $k$ respectively, the loss function is defined as follows:
\begin{equation*}
    \mathcal{L}_{drive} = \frac{1}{K} \sum_{k=1}^{K} \| \mathbf{w}_k^\star - \mathbf{w}_k^{gt} \|_1 
    \label{loss_cross}
\end{equation*}

\section{Environment}

\subsection{Driving on CARLA}
\subsubsection{Dataset}
Our dataset is built upon the CARLA~\cite{Dosovitskiy17} 0.9.10. 
As mentioned in Sec.~\ref{sec:ps}, we adopt an expert agent (the same as~\cite{Chitta2023PAMI}) in the CARLA environment to collect data along the routes defined in the Longest6~\cite{Jia2024NeurIPS} and LangAuto~\cite{shao2024lmdrive} benchmarks. Further, we augment the dataset with the semantic topology model from~\cite{shao2024lmdrive} to generate natural language navigation data, and the planning method from~\cite{ding2021epsilon} to generate planning corridor data. The corresponding BEVs are also collected following~\cite{Chitta2023PAMI}.
The whole dataset covers eight public towns in CARLA, with six towns (Town 1, 3, 4, 6, 7, 10) dedicated to training and two (Town 2, 5) for testing. 
It includes diverse driving scenarios with varied road structures, such as roundabouts, multi-lane highways, and complex intersections, and covers seven weather conditions (Clear, Cloudy, WetCloudy, SoftRain, MidRain, HardRain) and three daylight conditions (Noon, Night, Sunset).

Regarding the onboard sensors, we set up one LiDAR sensor, and four RGB cameras pointing to the front, left, right, and rear. The front camera is located at $(x:1.3, y:0.0, z:2.3)$ with a yaw angle of $0^{\circ}$, while the two side cameras are with the same location but angled at $-60^{\circ}$ and $60^{\circ}$, respectively. The rear camera is placed at $(x:-1.3, y:0.0, z:2.3)$ and angled at $180^{\circ}$. Each of the cameras outputs images with a resolution of $960\times480$ pixels, covering horizontal field of view (HFOV) of $120^{\circ}$.



\subsubsection{Metrics}
We conduct the driving testing for models on two CARLA benchmarks, LangAuto~\cite{shao2024lmdrive} and Longest6~\cite{Jia2024NeurIPS}. In each benchmark, a set of driving routes with specific settings is pre-defined. The driving agents are rolled out in the environment and evaluated the driving performance with metrics. For each driving route, we measure the driving performance of the agents with three metrics defined by CARLA Leaderboard~\cite{Dosovitskiy17}: the driving score (\textit{DS}), route completion (\textit{RC}), and infraction score (\textit{IS}), described below:
\begin{equation}
\left\{\begin{aligned}
    \textit{RC} &= \frac{L_{completed}}{L_{total}} \\
    \textit{IS} &= \textit{IS}_0 \times \prod_{n=1}^{N} \delta_n \\
    \textit{DS} &= \textit{RC} \times \textit{IS}
    \end{aligned}\right.
    \label{RC}
\end{equation}
where \(L_{total}\) and \(L_{completed}\) are the total route length, and the completed route length, respectively. \textit{RC} represents the percentage of the route distance completed by an agent;
\textit{IS} measures the traffic rule violations, where $\textit{IS}_0$ is the initial driving score. Regarding the occurred violation, $\textit{IS}_0$ is multiplied by the corresponding factor \(\delta_n\), and the scores of $N$ violations are summed up; \textit{DS} is the product of the \textit{RC} and the \textit{IS}.

By default setting of the CARLA Leaderboard, only a timeout or a deviation from the route can trigger the termination of the episode. We propose a new metric $\textit{RC}_\text{strict}$, 
\begin{equation*}
        \textit{RC}_\text{strict} = \frac{L_s}{L_{total}}.
    \label{RC_strict}
\end{equation*}
where $L_s$ denotes the route distance completed by the agent under zero tolerance for any traffic infraction, such as failing to stop at a red traffic light or colliding with vehicles/pedestrians.
The metric $\textit{RC}_\text{strict}$ reflects a higher restriction on driving safety.

We also use $\textit{M}_\text{prec}$ to measure the precision of the BEVs prediction. As defined below, the function in $GIoU(\cdot)$ computes the discrepancy of the predicted semantic BEVs, $\hat{\mathbf{B}}$ with ground-truth semantic BEVs, $\mathbf{B}$:
\begin{equation*}
        \textit{M}_\text{prec} = GIoU(\hat{\mathbf{B}}, Y) = \frac{\lvert \hat{\mathbf{B}} \cap \mathbf{B} \rvert}{\lvert \hat{\mathbf{B}} \cup \mathbf{B} \rvert} - \frac{\lvert \mathbf{C} \setminus (\hat{\mathbf{B}} \cup \mathbf{B}) \rvert}{\lvert \mathbf{C} \rvert},
    \label{Metric_giou}
\end{equation*}
where $\mathbf{C}$ is the area of the smallest enclosing convex shape in prediction and ground truth.

\subsection{VLM Fine-tuning with Corner Cases}
\subsubsection{Datasets}
We adopt the corner-case dataset CODA-LM~\cite{li2024automated} and DriveLM~\cite{sima2025drivelm} to fine-tune the VLM backbone of AppleVLM. CODA-LM comprises $9,768$ real-world driving scenarios (images), each attached with one question–answer (QA) pair generated by ChatGPT. In contrast, DriveLM introduces spatial-temporal reasoning with graph-structured VQAs from both nuScenes ($4,871$ frames) and CARLA ($183,000$ frames), covering sub-tasks of perception, prediction, and planning. 
To enable joint training by these two datasets, we annotated the DriveLM following the QA-pair format used in the CODA-LM to align data formats. For more details please refer to~\cite{li2024automated} and~\cite{sima2025drivelm}.

\subsubsection{Metrics}

For CODA-LM, we follow its benchmark~\cite{li2024automated} and report performance using the Text-Score (\textit{TS}) of correctness from GPT-4, normalized to $[1,100]$. For three tasks of CoT, we compute the $\textit{TS}_\text{gp}$, $\textit{TS}_\text{rp}$, and $\textit{TS}_\text{ds}$, and provide the $\textit{TS}_\text{avg}$ (the average of the three TS) in Tab.~\ref{tab_results_LLM}. For DriveLM, evaluation relies on \textit{SPICE}, \textit{GPT Score}, and \textit{Completeness} metrics. \textit{SPICE} converts predictions and ground truth into scene graphs and computes the \textit{F-score}. \textit{GPT Score} is a reasoning-based score from $0$ to $100$ judged by GPT-4. \textit{Completeness} measures the fraction of QAs correctly answered above a given threshold.

\begin{table*}[!t]
	\begin{center}
		\caption{Performance comparison of $8$ end-to-end driving models on the LangAuto \cite{shao2024lmdrive} benchmark. The $\uparrow$ stands for the higher the better. For each metric, the best result is shown in bold.}
		\label{tab_results_of_driving}
	    \setlength{\tabcolsep}{2pt}
		    \resizebox{\textwidth}{!}{\begin{tabular}{ccc|cccc|cccc|cccc}
        \toprule
            \multicolumn{2}{c}{Model} & Cond. & \multicolumn{4}{c}{LangAuto} & \multicolumn{4}{c}{LangAuto Short} & \multicolumn{4}{c}{LangAuto Tiny} \\
            \cmidrule{4-7} \cmidrule{8-11} \cmidrule{12-15}
            & & & $\textit{DS}$ $\uparrow$ & $\textit{RC}$ $\uparrow$ & $\textit{RC}_{\text{strict}}$ $\uparrow$ & $\textit{IS}$ $\uparrow$ 
			& $\textit{DS}$ $\uparrow$ & $\textit{RC}$ $\uparrow$ & $\textit{RC}_{\text{strict}}$ $\uparrow$ & $\textit{IS}$ $\uparrow$ 
			& $\textit{DS}$ $\uparrow$ & $\textit{RC}$ $\uparrow$ & $\textit{RC}_{\text{strict}}$ $\uparrow$ & $\textit{IS}$ $\uparrow$ \\ 
               \midrule
                \multicolumn{2}{c}{TCP~\cite{wu2022trajectoryguided}} & TP & $39.1 \pm 0.9$ & $73.5 \pm 2.2$ & $20.4 \pm 2.2$ & $0.46 \pm 0.06$ & $47.6 \pm 4.2$ & $78.1 \pm 3.1$ & $28.7 \pm 3.1$ & $0.61 \pm 0.03$ & $50.7 \pm 1.2$ & $89.1 \pm 4.3$ & $33.1 \pm 4.3$ & $0.57 \pm 0.06$ \\
              \multicolumn{2}{c}{Transfuser~\cite{Chitta2023PAMI}} & TP & $37.0 \pm 3.1$ & $74.0 \pm 3.5$ & $14.9 \pm 3.5$ & $0.50 \pm 0.06$ & $41.4 \pm 1.1$ & $81.2 \pm 2.6$ & $20.1 \pm 2.6$ & $0.51 \pm 0.03$ & $47.2 \pm 1.2$ & $93.0 \pm 5.0$ & $29.1 \pm 5.0$ & $0.50 \pm 0.06$ \\
            \multicolumn{2}{c}{Transfuser++ \cite{jaeger2023hidden}} & TP & $58.7 \pm 1.1$ & $\mathbf{82.1 \pm 2.5}$ & $56.2 \pm 3.0$ & $0.71 \pm 0.02$ & $64.8 \pm 1.9$ & $\mathbf{85.2 \pm 2.3}$ & $63.8 \pm 1.5$ & $0.76 \pm 0.04$ & $75.3 \pm 2.2$ & $\mathbf{95.5 \pm 2.3}$ & $74.1 \pm 1.2$ & $0.81 \pm 0.02$ \\
              \midrule
              \multicolumn{2}{c}{UniAD~\cite{hu2023planning}} & NC & $40.4 \pm 1.8$ & $80.0 \pm 2.9$ & $36.1 \pm 2.9$ & $0.51 \pm 0.04$ & $40.2 \pm 2.4$ & $83.1 \pm 4.4$ & $36.9 \pm 4.4$ & $0.61 \pm 0.02$ & $47.1 \pm 4.1$ & $89.1 \pm 1.1$ & $37.1 \pm 1.1$ & $0.75 \pm 0.03$ \\
              \multicolumn{2}{c}{Interfuser~\cite{shao2023safety}} & NC & $36.7 \pm 0.7$ & $72.1 \pm 4.1$ & $26.7 \pm 4.1$ & $0.51 \pm 0.06$ & $44.6 \pm 3.1$ & $74.4 \pm 4.6$ & $30.2 \pm 4.6$ & $0.60 \pm 0.08$ & $48.1 \pm 3.0$ & $74.5 \pm 5.0$ & $35.2 \pm 5.0$ & $0.65 \pm 0.05$ \\
              \multicolumn{2}{c}{LMDrive~\cite{shao2024lmdrive}} & NC & $36.2 \pm 2.3$ & $46.5 \pm 4.3$ & $41.1 \pm 4.3$ & $0.81 \pm 0.03$ & $50.6 \pm 1.7$ & $60.0 \pm 3.4$ & $50.6 \pm 1.7$ & $0.84 \pm 0.04$ & $66.5 \pm 3.6$ & $77.9 \pm 2.3$ & $54.8 \pm 2.3$ & $0.85 \pm 0.02$ \\
            \midrule
            \multirow{2}{*}{AppleVLM} &  LLaVA & NC & $52.5 \pm 1.5$ & $63.2 \pm 1.3$ & $56.2 \pm 1.5$ & $0.83 \pm 0.04$ & $59.9 \pm 1.1$ & $71.3 \pm 1.5$ & $57.3 \pm 1.4$ & $0.84 \pm 0.02$ & $67.2 \pm 1.6$ & $78.1 \pm 1.2$ & $68.7 \pm 1.1$ & $0.86 \pm 0.02$ \\
            & Janus Pro & NC & $\mathbf{59.2 \pm 1.9}$ & $67.1 \pm 3.4$ & $\mathbf{59.1 \pm 3.1}$ & $\mathbf{0.89 \pm 0.04}$ & $\mathbf{66.0 \pm 1.6}$ & $76.9 \pm 3.7$ & $\mathbf{64.5 \pm 3.9}$ & $\mathbf{0.91 \pm 0.03}$ & $\mathbf{76.0 \pm 3.3}$ & $83.5 \pm 2.0$ & $\mathbf{76.9 \pm 1.9}$ & $\mathbf{0.91 \pm 0.02}$ \\
            \bottomrule
    \end{tabular}}
	\end{center}
\end{table*}

\begin{table}[!t]
	\begin{center}
        \fontsize{5.3pt}{7pt}\normalfont
		\caption{Performance comparison of $8$ end-to-end driving models on Longest6 \cite{Jia2024NeurIPS} benchmark. The $\uparrow$ stands for the higher the better. For each metric, the best result is shown in bold.}
		\label{tab_results_of_driving_2}
        \scriptsize
        \setlength{\tabcolsep}{1pt}
        \begin{tabular}{ccc cccc}
        			\toprule
        			\multicolumn{2}{c}{Model} & Cond. & $\textit{DS}$ $\uparrow$ & $\textit{RC}$ $\uparrow$ & $\textit{RC}_{\text{strict}}$ $\uparrow$ & $\textit{IS}$ $\uparrow$ \\ 
        			\midrule
                    \multicolumn{2}{c}{TCP~\cite{wu2022trajectoryguided}} & TP & $32.5 \pm 0.9$ & $81.2 \pm 1.2$  & $17.2 \pm 2.2$ & $0.40 \pm 0.05$ \\
                    \multicolumn{2}{c}{Transfuser~\cite{Chitta2023PAMI}} & TP & $38.6 \pm 0.9$ & $84.1 \pm 1.2$  & $15.3 \pm 3.5$ & $0.42 \pm 0.06$ \\
                    \multicolumn{2}{c}{Transfuser++~\cite{jaeger2023hidden}} & TP & $\mathbf{62.3 \pm 1.1}$ & $\mathbf{89.0 \pm 1.0}$  & $59.2 \pm 3.2$ & $0.70 \pm 0.01$ \\
                    \midrule
                    \multicolumn{2}{c}{UniAD~\cite{hu2023planning}} & NC & $38.1 \pm 1.8$ & $79.9 \pm 1.7$  & $34.2 \pm 2.9$ & $0.46 \pm 0.06$ \\
                    \multicolumn{2}{c}{Interfuser~\cite{shao2023safety}} & NC & $45.4 \pm 0.9$ & $88.9 \pm 5.3$ & $18.9 \pm 4.1$ & $0.51 \pm 0.06$ \\
                    \multicolumn{2}{c}{LMDrive~\cite{shao2024lmdrive}} & NC & $35.8 \pm 2.3$ & $43.1 \pm 2.5$  & $38.1 \pm 4.3$ & $0.83 \pm 0.03$ \\
                    \midrule
        			\multirow{2}{*}{AppleVLM} & LLaVA & NC & $53.4 \pm 3.1$ & $62.1 \pm 2.7$  & $54.2 \pm 4.3$ & $0.86 \pm 0.02$ \\
                    & Janus Pro & NC & $60.5 \pm 2.1$ & $68.7 \pm 3.7$ & $\mathbf{60.9 \pm 1.2}$ & $\mathbf{0.88 \pm 0.02}$ \\
        			\bottomrule
        \end{tabular}
	\end{center}
\end{table}

\section{Experiments}

\subsection{Experimental Setup}
\subsubsection{Training Details}
We took RegNet64 \cite{9743274} as the backbone to process the RGB images and LiDAR point-cloud data in the vision encoder and Janus Pro \cite{chen2025janus} as the backbone of our VLM decoder. All three stages of training were with the AdamW optimizer \cite{loshchilov2018fixing}. When pre-training the vision encoder with BEV maps,
we used an initial learning rate (LR) of $5e^{-5}$ and weight decay of $0.01$. We trained the model for $50$ epochs on one NVIDIA $4090$ GPU, with a batch size of $64$. The LR was scheduled with cosine annealing \cite{loshchilov2017}, which dynamically adjusted LR in the shape of a cosine curve until the minimum LR=$0$.
During the fine-tuning of the VLM decoder, we set an initial learning rate of $2e^{-5}$ and weight decay of $0.05$. The model was trained for $12$ epochs on $10$ NVIDIA A6000 GPUs in parallel, with a batch size of $12$. The LR decays following cosine annealing.
To train the end-to-end AppleVLM, we used an initial learning rate of $1e^{-5}$ and weight decay of $0.01$. We trained it for $10$ epochs on $10$ NVIDIA A6000 GPUs in parallel, with a batch size of $8$. The learning rate linearly decayed with warm-up in the first $5\%$ of the training steps.

\subsubsection{Real-world Deployment}
\begin{figure}
    \centering
    \includegraphics[width=1\linewidth]{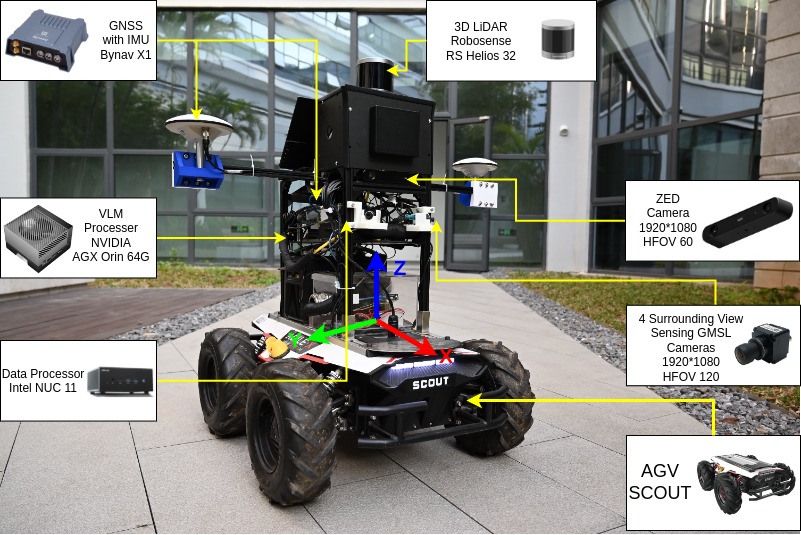}
    \caption{AGV platform for real-world deployment}
    \label{figure_deployment}
\end{figure}

We deployed the learned end-to-end driving policy of AppleVLM on a Scout automated guided vehicle(AGV) platform. As depicted in Fig.~\ref{figure_deployment}, the AGV was equipped with an NVIDIA Jetson AGX Orin with 64GB SOC processor for VLM process and Intel NUC 11 for data process. The sensor suite consisted of a Bynav X1 GNSS and IMU for approximate localization and pose, a Robosense RS Helios 32 LiDAR sensor, one ZED camera with $60^{\circ}$ HFOV at resolution of $1920 \times 1080$, and four Sensing SG2-GMSL cameras with $120^{\circ}$ HFOV at resolution of $1920 \times 1080$ pixels for capturing visual observations. The navigation instruction and global navigation information came from Amap navigation SDK.

\begin{figure*}[!t]
    \centering
    \begin{subfigure}[b]{0.46\textwidth}
        \includegraphics[width=\linewidth]{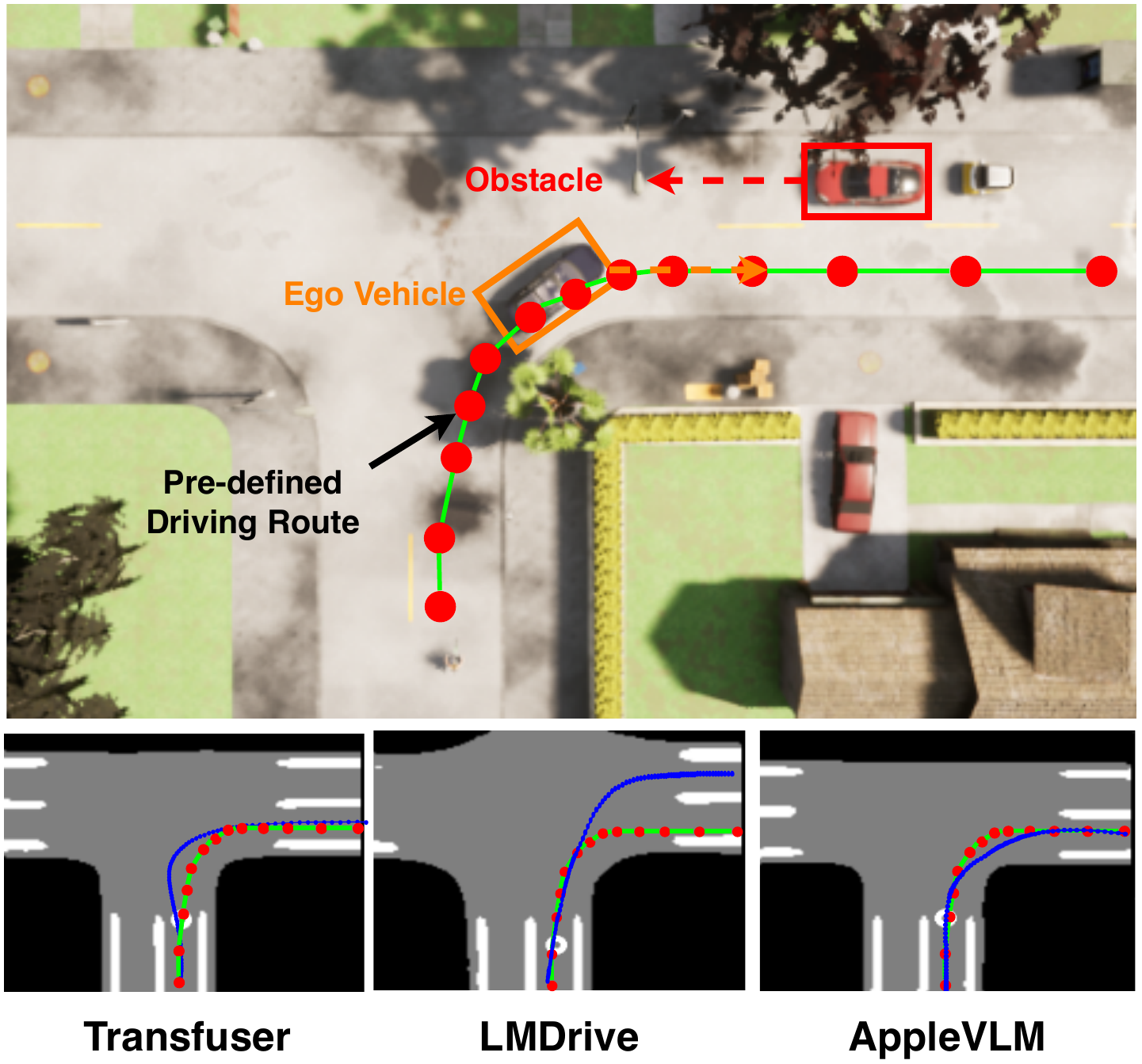}
        \scriptsize
        \caption{Normal right turning}
        \label{fig:subfig-carlas1}
    \end{subfigure}
    \hfill
    \begin{subfigure}[b]{0.46\textwidth}
        \includegraphics[width=\linewidth]{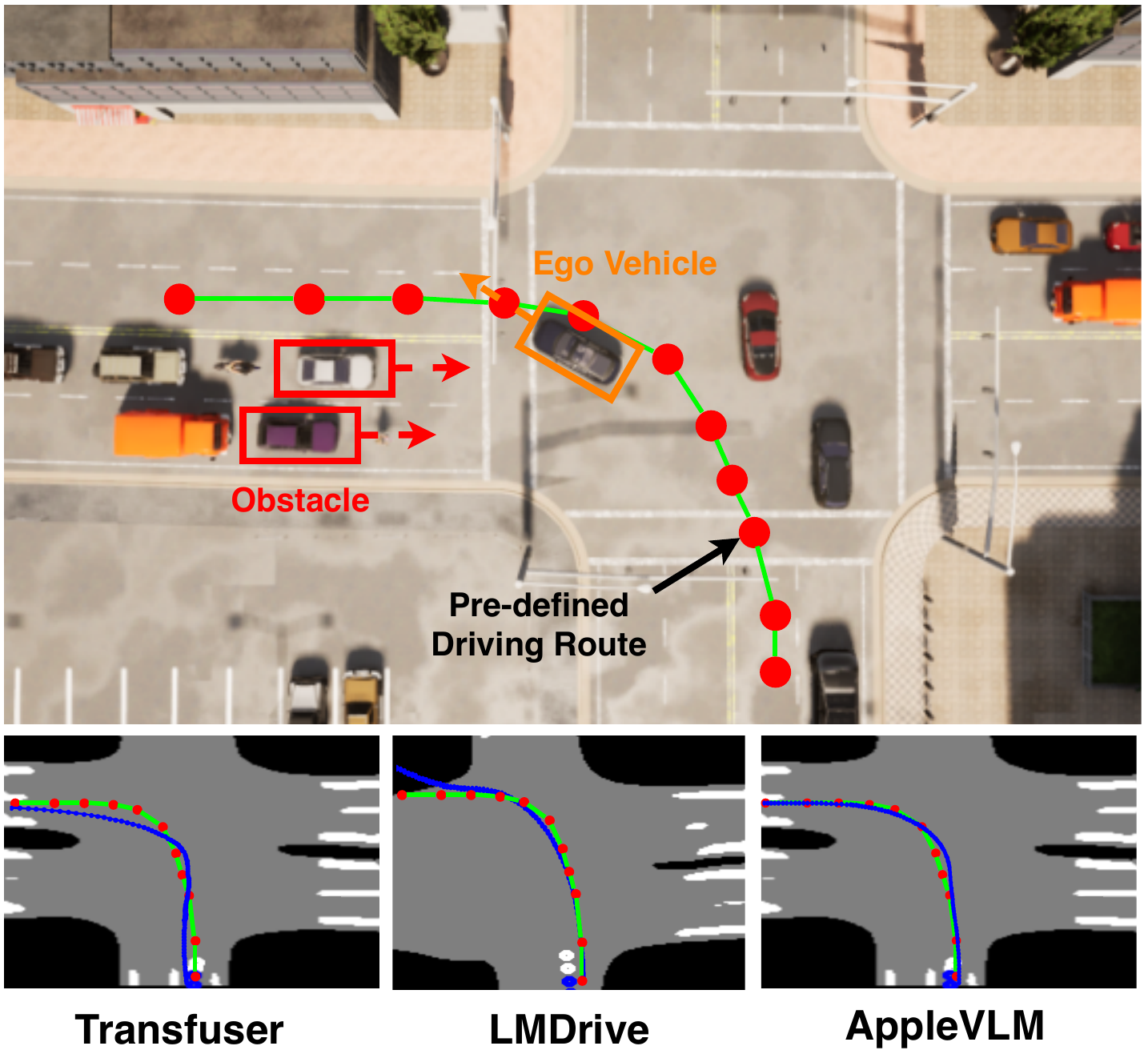}
        \scriptsize
        
        \caption{Unprotected left turning}
        \label{fig:subfig-carlas2}
    \end{subfigure}
    
    \begin{subfigure}[b]{0.46\textwidth}
        \includegraphics[width=\linewidth]{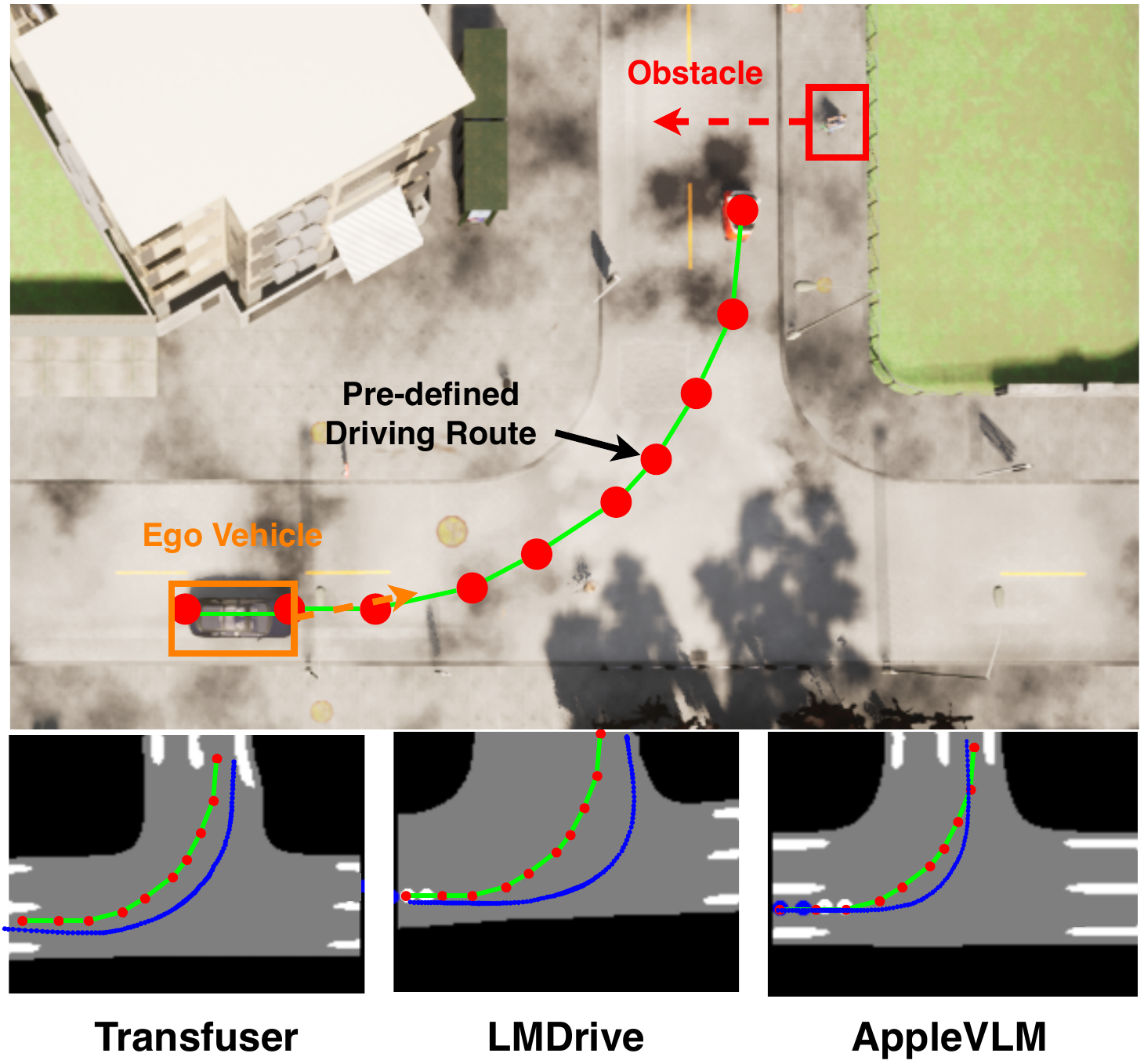}
        
        \scriptsize
        \caption{Left turning with pedestrian intrusion modeled by game theory}
        \label{fig:subfig-carlas3}
    \end{subfigure}
    \hfill
    \begin{subfigure}[b]{0.46\textwidth}
        \includegraphics[width=\linewidth]{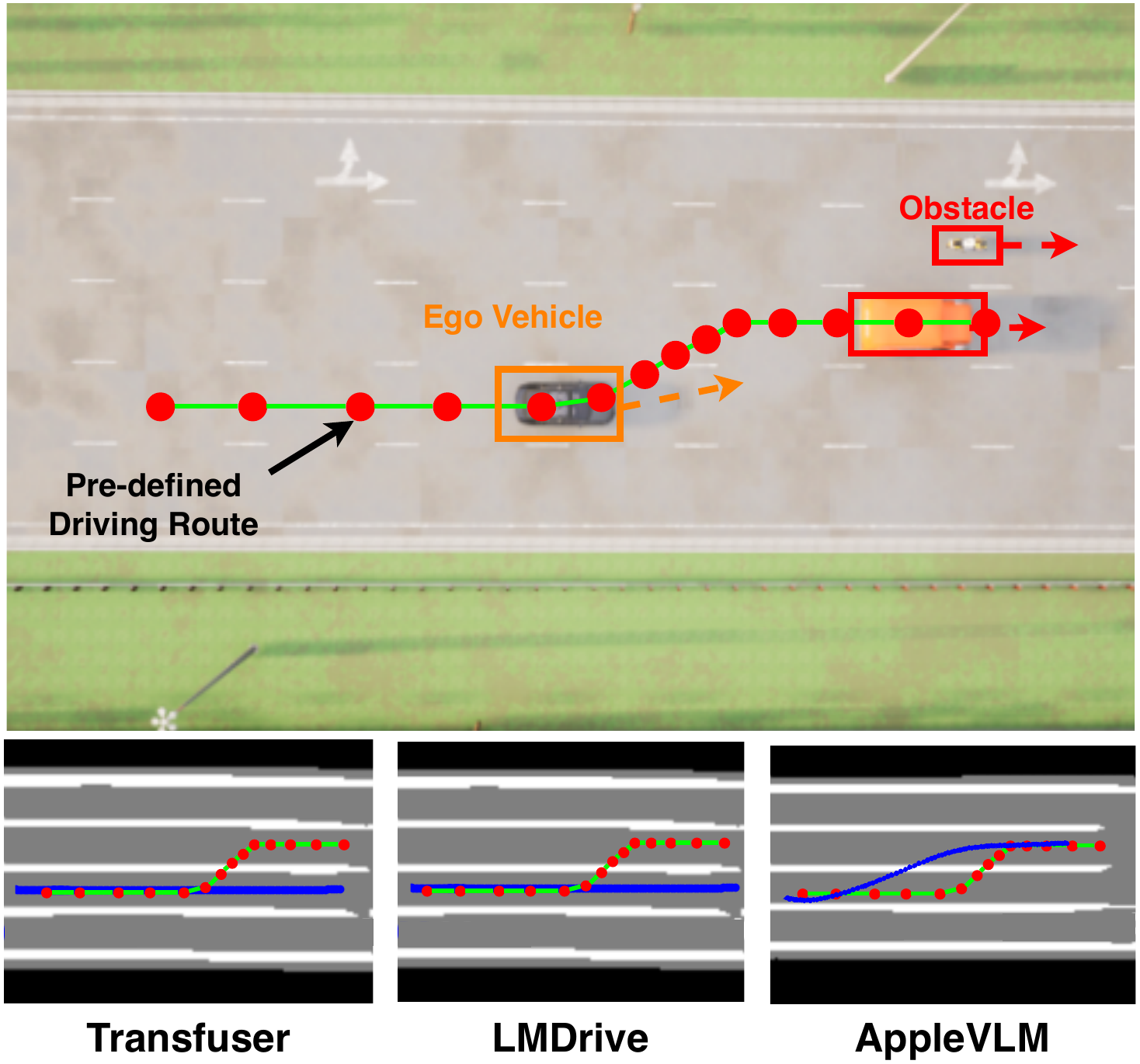}

        \scriptsize
        \caption{Lane changes with dynamic obstacles}
        \label{fig:subfig-carlas4}
    \end{subfigure}
    \caption{Qualitative results of Transfuser, LMDrive, and AppleVLM in four specific driving scenarios. The driving trajectories are completed in a closed-loop driving environment on the CARLA simulator. In each sub-figure, we show on top the driving scenario from a bird's eye view. The orange block along with a pointer indicates the ego vehicle and its driving direction. The red dotted line represents the dynamic obstacles that potentially affect the ego vehicle, and the pointer points to its driving direction. At the bottom, we show the predicted BEVs with driving trajectories. The green lines with red dots indicate the pre-defined driving trajectories, and the blue ones are the actual driving trajectories.}
    \label{fig_simulation}
\end{figure*}

\subsection{CARLA Experiments}
The experiments on the CARLA simulator consist of four parts. Firstly, we perform the driving test for all models on two public benchmarks (LangAuto \cite{shao2024lmdrive} and Longest6 \cite{Chitta2023PAMI}) to provide quantitative results and analysis. Then, for intuitive comparison, we qualitatively evaluate the model's performance in four specific scenarios defined in our study: normal right turn, unprotected left turn, unprotected left turn with pedestrian intrusions modeled by game theory, and lane change with dynamic obstacles.
Next, we compare the influence of changes in sensor settings and show the robustness of AppleVLM. Finally, we conduct an ablation study on key parts of the AppleVLM.

\subsubsection{Quantitative Results}
In Tab.~\ref{tab_results_of_driving} and \ref{tab_results_of_driving_2}, we summarize the quantitative results of different methods tested on LangAuto and Longest6 benchmarks. 
For comparison, we select five typical uni-modal end-to-end driving models that are solely based on vision: Transfuser~\cite{Chitta2023PAMI}, Interfuser~\cite{shao2023safety}, TCP~\cite{wu2022trajectoryguided}, UniAD~\cite{hu2023planning} and Transfuser++~\cite{jaeger2023hidden}. Since there are few existing VLM-based driving models, we only consider the latest proposed LMdrive~\cite{shao2024lmdrive}, along with two AppleVLM models embedded with the LLaVA v1.5~\cite{liu2024llavanext} and Janus Pro~\cite{chen2025janus} backbones, fine-tuned with both CODA-LM and DriveLM datasets. 
As introduced in Transfuser++~\cite{jaeger2023hidden}, in the table, we use \textit{TP} (map-based GNSS localization of the target points) and \textit{NC} (discrete commands \eg, follow lane, turn right) to indicate models that receive navigation information based on two different conditions (Cond.). Considering the variance in different training runs, we use $3$ different random seeds for the training of each method. For each method, we select the model with the best offline evaluation results (\ie, the lowest mean absolute error) and further perform the driving test on CARLA benchmarks. To account for the non-determinism of closed-loop driving on CARLA, we repeat the driving test of each model $3$ times and report their mean and standard deviation in the table.

It can be found from Tab.~\ref{tab_results_of_driving} that the proposed AppleVLM achieves the best performance in both \textit{DS} and \textit{IS}among three sub-tasks pf the LangAuto benchmark. Although the \textit{TP} conditioned TransFuser++ outperforms AppleVLM in \textit{RC}, it conditions predictions using the next target point along the route, which tends to be more easily recover from steering errors (proved in~\cite{jaeger2023hidden}); in contrast, AppleVLM only uses the discrete navigation commands that contain no geometric information about the center of the lanes.
Compared to the latest proposed VLM-based method LMDrive, AppleVLM improves \textit{RC} by $16.7$ and $20.6$ in long-distance driving (longer than $500$ meters). In two short-distance driving tasks that contain routes less than $500$ meters, AppleVLM still maintains the best performance, leading by $11.3$ and $16.9$ in LangAuto short, while $0.2$ and $5.6$  in LangAuto Tiny, respectively. 

Compared to the vision-based models, the VLM-based models generally obtain lower \textit{RC} results. We observe that this is because in the original benchmark setting, an episode is not terminated when a collision occurs unless the ego vehicle deviates from the desired lane.
Inspired by \cite{10341506}, we propose the $\textit{RC}_\text{strict}$ metric, where an episode is terminated and counted as failed once any traffic violation occurs (including collision, route deviation {\etc}). In this case, we find that the vision-based models have a significant decline in completing the route driving, and all the VLM-based models can achieve better results than the vision-based models. AppleVLM with fine-tuned Janus pro achieves the best $\textit{RC}_\text{strict}$ results among all models with the lead of $18.0$, $13.9$ and $22.1$ compared to the LMDrive, and $2.9$, $0.7$ and $2.8$ to the Transfuser++. Moreover, we can find that the results of Longest6 (Tab.~\ref{tab_results_of_driving_2}) demonstrate the same conclusion as LangAuto, which further shows the remarkable advantage of AppleVLM over other models.

\begin{table*}[!t]
	\begin{center}
		\caption{The influence of changes in sensor configuration for all models. The models are tested on LangAuto~\cite{shao2024lmdrive} benchmark. The $\uparrow$ stands for the higher the better.}
		\label{tab_results_different_sensors_deployment}
        \setlength{\tabcolsep}{3pt}
        \resizebox{\textwidth}{!}{\begin{tabular}{
            c|cc|cc|cc|cc|cc|cc|cc|cc}
			\toprule 
             & \multicolumn{4}{c|}{Transfuser}& \multicolumn{4}{c|}{LMDrive} & \multicolumn{4}{c|}{AppleVLM w/o $\mathcal{DA}$} & \multicolumn{4}{c}{AppleVLM with $\mathcal{DA}$} \\
		Setting & $\textit{M}_\text{prec}$ $\uparrow$ & Gain  
                & $\textit{RC}$ $\uparrow$  & Gain 
                & $\textit{M}_\text{prec}$ $\uparrow$ & Gain  
                & $\textit{RC}$ $\uparrow$ & Gain 
                & $\textit{M}_\text{prec}$ $\uparrow$ & Gain  
                & $\textit{RC}$ $\uparrow$ & Gain 
                & $\textit{M}_\text{prec}$ $\uparrow$ & Gain  
                & $\textit{RC}$ $\uparrow$ & Gain \\
            \midrule
             Original 
            & $69.11$ & - & $74.0 \pm 3.5$ & - 
            & $63.27$ & - & $46.5 \pm 4.3$ & - 
            & $67.23$ & - & $59.9 \pm 1.7$ & - 
            & $71.12$ & - & $67.1 \pm 3.4$ & -\\
            
            $1$ & $47.11$ & $-22.00$ & $21.5 \pm 4.3$ & $-52.5$ & $43.56$ & $-19.71$ & $23.0 \pm 3.5$ & $-23.5$ 
            & $53.96$ & $-13.27$ & $54.2 \pm 3.1$ & $-5.7$
            & $60.55$ & $-10.57$ & $63.6 \pm 3.9$ & $-3.5$\\
            
            $2$ & $31.02$ & $-38.09$ & $17.3 \pm 1.7$ & $-56.7$ & $34.70$ & $-28.57$ & $18.7\pm 3.2$ & $-27.8$ 
            & $50.73$ & $-16.50$ & $47.6 \pm 2.5$ & $-12.3$ 
            & $57.12$ & $-14.00$ & $58.2 \pm 3.7$ & $-8.9$\\
            
            $3$ & $28.21$ & $-38.09$ & $15.2\pm 2.3$ & $-58.8$ & $34.10$ & $-29.17$ & $15.8 \pm 2.9$ & $-30.7$ 
            & $45.52$ & $-21.71$ & $44.1 \pm 2.3$ & $-15.8$
            & $53.09$ & $-18.03$ & $55.6 \pm 3.2$ & $-11.5$\\
            
            $4$ & $27.33$ & $-41.78$ & $27.2\pm 2.9$ & $-46.8$ & $30.94$ & $-32.33$ & $14.4 \pm 0.5$ & $-32.1$ 
            & $42.78$ & $-24.45$ & $51.6 \pm 2.9$ & $-8.3$ 
            & $52.17$ & $-18.95$ & $63.0 \pm 1.4$ & $-4.1$\\
            
            $5$ & $51.55$ & $-17.56$ & 
            $13.4\pm 1.5$ & $-60.6$& $44.98$ & $-18.29$ & $21.0 \pm 1.7$ & $-25.5$ 
            & $50.10$ & $-17.13$ & $38.4 \pm 1.3$ & $-21.5$
            & $54.51$ & $-16.61$ & $51.4 \pm 0.6$ & $-15.7$\\
		\bottomrule
		\end{tabular}}

	\end{center}
    \label{table:sensorconfig}
    	\footnotesize{
		\textbf{Note:} Parameters different from the original setting are shown \textbf{in bold}: \\
        \\
        \begin{minipage}{\textwidth}
        \centering
        \begin{tabular}{c | c c | c c | c c }
        \toprule
        \multirow{2}{*}{Setting} & \multicolumn{2}{c|}{Front} & \multicolumn{2}{c|}{Left} & \multicolumn{2}{c}{Right} \\
                & Position & Rotation & Position & Rotation & Position & Rotation \\
                \midrule
         Original & $(1.3, 0.0, 2.3)$ & $(0^\circ, 0^\circ, 0^\circ)$ & $(1.3, 0.0, 2.3)$ & $(0^\circ, 0^\circ, -60^\circ)$ & $(1.3, 0.0, 2.3)$ & $(0^\circ, 0^\circ, 60^\circ)$ 
         \\  
         $1$ & $(1.3, 0.0, 2.3)$ & $(0^\circ, 0^\circ, 0^\circ)$ & $(1.3, 0.0, 2.3)$ & $(0^\circ, 0^\circ, -\mathbf{90}^\circ)$ & $(1.3, 0.0, 2.3)$ & $(0^\circ, 0^\circ, \mathbf{90}^\circ)$ 
         \\   
         $2$ & $(1.3, 0.0, 2.3)$ & $(0^\circ, 0^\circ, 0^\circ)$ & $\mathbf{(1.3, 0.5, 2.3)}$ & $(0^\circ,0^\circ, -\mathbf{90}^\circ)$ & $\mathbf{(1.3, 0.5, 2.3)}$ & $(0^\circ, 0^\circ, \mathbf{90}^\circ)$ 
         \\  
         $3$ & $(\mathbf{0.8}, 0.0, 2.3)$ & $(0^\circ, 0^\circ, 0^\circ)$ & $(1.3, 0.0, 2.3)$ & $(0^\circ, 0^\circ, -60^\circ)$ & $(1.3, 0.0, 2.3)$ & $(0^\circ, 0^\circ, 60^\circ)$ 
         \\
         $4$ & $(1.3, 0.0, 2.3)$ & $(0^\circ, \mathbf{-30}^\circ, 0^\circ)$ & $(1.3, 0.0, 2.3)$ & $(0^\circ, 0^\circ, -60^\circ)$ & $(1.3, 0.0, 2.3)$ & $(0^\circ, 0^\circ, 60^\circ)$ 
         \\
         $5$ & $(1.3, 0.0, 2.3)$ & $(\mathbf{30}^\circ, 0^\circ, 0^\circ)$ & $(1.3, 0.0, 2.3)$ & $(0^\circ, 0^\circ, -60^\circ)$ & $(1.3, 0.0, 2.3)$ & $(0^\circ, 0^\circ, 60^\circ)$ 
         \\
         \bottomrule
        \end{tabular}
        \end{minipage}
	}
\end{table*}

\subsubsection{Qualitative Analysis}
Fig.~\ref{fig_simulation} shows the driving trajectories of the vision-based driving model Transfuser and the VLM-based driving model LMDrive, and the proposed AppleVLM across four specific scenarios, depicted in a BEV format. In general, AppleVLM consistently generates driving trajectories that most closely align with the desired ones in all scenarios. Specifically, LMDrive is prone to deviating from the driving lane (either driving into the opposite lane or the sidewalk) when turning (see Fig.~\ref{fig:subfig-carlas1} and \ref{fig:subfig-carlas2}). Transfuser can mostly complete the turning but tends to drift unnecessarily (see Fig.~\ref{fig:subfig-carlas1} and \ref{fig:subfig-carlas2}). When a lane change is required, only AppleVLM can respond effectively and make a reasonable lane change in the presence of other vehicles (see Fig.~\ref{fig:subfig-carlas4}).

\subsubsection{Robust to Sensor Configuration Changes}

To examine the robustness of AppleVLM to the sensor change, we conduct experiments using Transfuser, LMDrive and AppleVLM without $\mathcal{DA}$ as the comparative models. We alter the position $(x, y, z)$ and rotation (roll, pitch, yaw) of three cameras and make 5  sensor settings different from the original setting during data collection, as detailed in Tab.~\ref{tab_results_different_sensors_deployment}. The comparison experiments are conducted on the LangAuto benchmark. We use \textit{RC} as the metric to reflect the driving performance. In addition, we provide $\textit{M}_\text{prec}$ of BEV prediction, which roughly reflects the interpretability of the visual encoder, as a reference.

Comparing AppleVLM to Transfuser and LMDrive, we see that under setting $1$ and $2$, Transfuser is extremely sensitive to the change in two side cameras, which is reflected in the rapid decline of $-52.5$ and $-56.7$ in \textit{RC}. LMDrive performs slightly better than Transfuser, with only half of the performance drop ($-23.5$ and $-27.8$ in \textit{RC}). In contrast, the AppleVLM is more robust with only slight decreases of $-3.5$ and $-8.9$. A similar trend is observed in the prediction of BEVs, where AppleVLM outperforms Transfuser and LMDrive with the smaller decrease in $\textit{M}_\text{prec}$ results. All models are affected by the position change in the front camera (setting $3$), while the Transfuser and LMDrive exhibit a dramatic performance decline in two metrics ($-38.09$ and $-58.8$, $-29.17$ and $-30.7$) compared to AppleVLM ($-18.03$ and $-11.5$). When adjusting the pitch of the front camera to $-30^\circ$ (setting $4$), the \textit{RC} results of Transfuser and LMDrive decreased by $46.8$ and $32.1$ respectively, approximately $10$ and $8$ times the decline observed in AppleVLM; and the BEV prediction performance of all models declined, with Transfuser exhibiting a dramatic drop of $-41.78$. A similar trend is observed in setting $5$, when changing the roll to $30^\circ$, the \textit{RC} of Transfuser is significantly reduced by $60.6$, which is almost $5$ times the decline of AppleVLM. Although LMDrive performs more stably than Transfuser under this setting, it still drops nearly $10$ points more than AppleVLM. Notably, through comparative experiments on whether to apply $\mathcal{DA}$ to AppleVLM, we observe that for all sensor changes, AppleVLM with $\mathcal{DA}$ consistently exhibits less decline in both $\textit{M}_\text{prec}$ and \textit{RC} metrics. This demonstrates that $\mathcal{DA}$ plays a crucial role in helping AppleVLM to maintain more stable performance when sensor conditions change.

\subsubsection{Ablation Study}
In the ablation study, we are interested in the impact of several key parts of AppleVLM: the proposed novel vision encoder, the planning strategy encoder for multi-modal feature fusion, and the corner-case fine-tuning for the VLM backbone.

\begin{table}[!t]
	\begin{center}
    
        \fontsize{7pt}{9pt}\normalfont  
		\caption{Ablation study on the impact of vision encoder design. The models are tested on LangAuto \cite{shao2024lmdrive} benchmark. The $\uparrow$ stands for the higher the better. For each metric, the best result is shown in bold.
        }
		\label{tab_results_module_design} 
        \setlength{\tabcolsep}{3.5pt}
		\begin{tabular}{cc c ccc}
		\toprule
		\multicolumn{2}{c}{Vision Encoder} & $\textit{M}_\text{precision}$ $\uparrow$&  $\textit{DS}$ $\uparrow$ & $\textit{RC}$ $\uparrow$ & $\textit{IS}$ $\uparrow$  \\
			\midrule
            \multicolumn{2}{c}{LMDrive \cite{shao2024lmdrive}} & $47.23$ & $29.8 \pm 2.3$ & $46.5 \pm 4.3$ & $0.64 \pm 0.03$ \\
		\multicolumn{2}{c}{Transfuser \cite{Chitta2023PAMI}} & $69.11$ & $33.1 \pm 3.1$ & $51.0 \pm 3.5$ & $0.66 \pm 0.06$ \\
            \multicolumn{2}{c}{BEVformer \cite{li2022bevformer}} & $57.82$ & $37.6 \pm 1.5$ & $53.0 \pm 4.3$ & $0.71 \pm 0.03$ \\
            \midrule
            
            \multirow{3}{*}{AppleVLM}&ResNet-50 & $59.91$ & $42.5 \pm 1.8$ & $55.3 \pm 3.0$ & $0.75 \pm 0.01$ \\
		&RegNet-32 & $65.79$ & $48.6 \pm 1.1$ & $60.1 \pm 1.2$ & $0.81 \pm 0.02$ \\
		&RegNet-64 & $\mathbf{71.12}$ & $\mathbf{52.6 \pm 2.2}$ & $\mathbf{65.5 \pm 5.3}$ & $\mathbf{0.88 \pm 0.04}$ \\
			\bottomrule
		\end{tabular}
	\end{center}
\end{table}

\noindent \textbf{Vision Encoder:}
We replace our proposed vision encoder in AppleVLM with the ones from LMDrive~\cite{shao2024lmdrive}, Transfuser~\cite{Chitta2023PAMI}, and BEVformer~\cite{li2022bevformer} respectively to verify its efficiency. In Tab.~\ref{tab_results_module_design}, we provide results of $\textit{M}_\text{precision}$ that reflect on the lane perception and \textit{DS}, \textit{RC} and \textit{IS} that indicate the driving performance. The results illustrate that AppleVLM embedded with our proposed vision encoder achieves the highest scores in all metrics, whereas the one with the ResNet-64 backbone yields the best performance. Meanwhile, changing the ResNet backbone might slightly alter the driving results, while all of them are superior to the above models. This fully reveals the effectiveness of our vision encoder based on the deformable attention.

\begin{table}[!t]
	\begin{center}
		\caption{Ablation study on the impact of planning strategy encoder. The models are tested on LangAuto \cite{shao2024lmdrive} benchmark. The $\uparrow$ stands for the higher the better. For each metric, the best result is shown in bold.}
		\label{tab_results_planning_template} 
        \setlength{\tabcolsep}{3pt}
		\begin{tabular}{cc | ccc}
		\toprule   
            \multicolumn{2}{c}{Model} &  $\textit{DS}$ $\uparrow$ & $\textit{RC}$ $\uparrow$ & $\textit{IS}$ $\uparrow$ \\
	     \midrule
            \multirow{2}{*}{LLaVA v1.5} & w/o $\mathbf{\mathcal{T}}_{P}$ & $36.2 \pm 2.3$ & $46.5 \pm 4.3$ & $0.81 \pm 0.03$ \\
            & w/ $\mathbf{\mathcal{T}}_{P}$ & $44.6 \pm 1.1$ & $57.2 \pm 5.3$ & $0.78 \pm 0.01$ \\
            \midrule
            \multirow{2}{*}{Janus Pro} & w/o $\mathbf{\mathcal{T}}_{P}$ & $40.4 \pm 1.7$ & $50.5 \pm 1.6$ & $0.80 \pm 0.03$ \\
            & w/ $\mathbf{\mathcal{T}}_{P}$ & $48.7 \pm 2.3$ & $60.1 \pm 3.1$ & $0.81 \pm 0.02$ \\
            \midrule
            Fine-tuned & w/o $\mathbf{\mathcal{T}}_{P}$  & $44.1 \pm 2.2$ & $51.1 \pm 5.3$ & $0.88 \pm 0.04$ \\
            Janus Pro & w/ $\mathbf{\mathcal{T}}_{P}$  & $\mathbf{52.6 \pm 2.2}$ & $\mathbf{65.5 \pm 5.3}$ & $\mathbf{0.88 \pm 0.04}$ \\
			\bottomrule
		\end{tabular}
	\end{center}
\end{table}

\begin{figure*}[!t]
    \begin{center}

    \begin{minipage}[b]{0.32\linewidth}
        \centering
        \setlength{\fboxrule}{2pt}   
        \setlength{\fboxsep}{0pt}    
        \includegraphics[width=\linewidth]{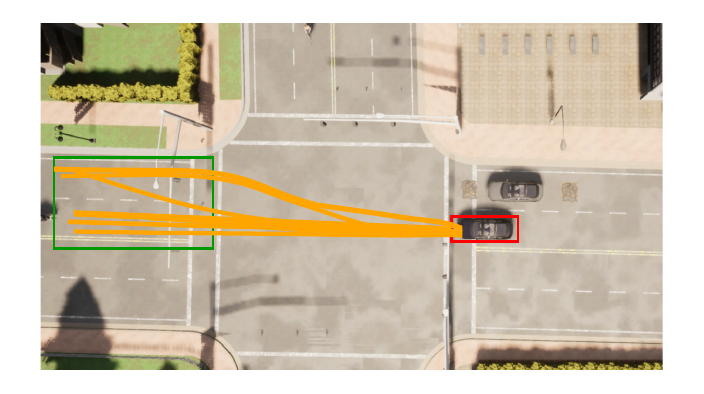} 
        
        \small 
        a) Command: ``Go straight at intersection", without PSE.
    \end{minipage}
    \hfill
    \begin{minipage}[b]{0.32\linewidth}
        \centering
        \setlength{\fboxrule}{2pt}   
        \setlength{\fboxsep}{0pt}    
        \includegraphics[width=\linewidth]{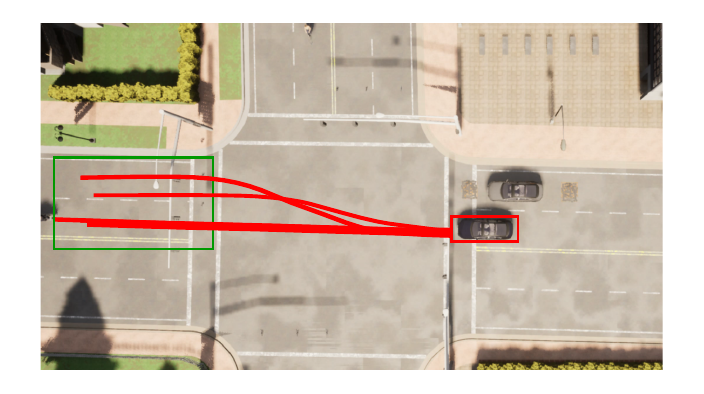} 
        
        \small
        b) Command: ``Go straight at intersection, select left lane", without PSE.
    \end{minipage}
    \hfill
    \begin{minipage}[b]{0.32\linewidth}
        \centering
        \setlength{\fboxrule}{2pt}   
        \setlength{\fboxsep}{0pt}    
        \includegraphics[width=\linewidth]{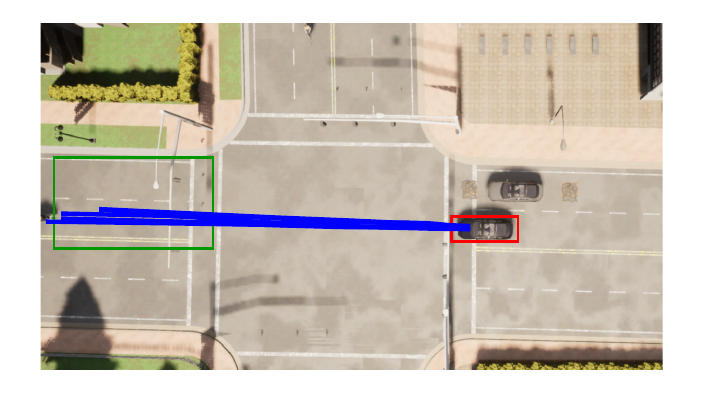} 
        
        \small
        c) Command: ``Go straight at intersection", with PSE.
    \end{minipage}

    \vspace{1ex}
    \caption{Qualitative analysis of the effectiveness of the planning strategy encoder in eliminating lane oscillations when transitioning from an intersection to a multi-lane road. Each condition was evaluated over $10$ runs.}
    \label{fig:three_subfigs}
\end{center}
\end{figure*}

\noindent \textbf{Planning-enhanced VLM:}
As described in Fig.~\ref{fig_Qformer}, compared to the traditional VLMs, AppleVLM incorporates a planning strategy encoder module, aiming to enhance the feature extraction and fusion for better driving performance.
To verify this, we conduct three sets of comparative experiments: the vanilla LLaVA v1.5, the vanilla Janus Pro, and the fine-tuned Janus Pro, all with and without a planning strategy encoder module. The driving results on the LangAuto benchmark are summarized in Tab.~\ref{tab_results_planning_template}.  In general, incorporating the planning features into the Q-former feature fusion improves \textit{DS} and \textit{RC} results for all models, where the fine-tuned Janus Pro one shows the most significant increase ($+8.5$ and $+14.4$). During the driving tests, we observe that when the driving model receives navigation commands in the form of natural language, it often understands the navigation language commands incorrectly and hence responds incorrectly, resulting in the ego vehicle deviating from the desired driving route. An example is shown in Fig.~\ref{fig_simulation}(d).
The planning strategy encoder provides explicit spatial information for navigation, thus incorporating it facilitates feature fusion of the vision and language modalities. To further explore how language input and the Planning Strategy Encoder (PSE) affect the model performance of following instructions, in Fig.~\ref{fig:three_subfigs}, we provide a scenario in which the ego vehicle passes by an intersection and drives into a multi-lane road under different conditions: a basic or more precise navigation command, with or without PSE.
The results reveal that the navigation ambiguity arises from language interpretation, while PSE can effectively solve the lane-oscillation problem, leveraging the explicit spatial information from planning tokens.

\begin{table}[!t]
	\begin{center}
        \fontsize{8pt}{8pt}\normalfont  
		\caption{Ablation study on the impact of corner-case fine-tuning for text prediction of VLMs. The models are tested on the CODA-LM~\cite{li2024automated} benchmark. The $\uparrow$ stands for the higher the better. For each metric, the best result is shown in bold.}
		\label{tab_results_LLM}  
        \setlength{\tabcolsep}{3pt}
		\begin{tabular}{cc cccc}
			\toprule 
			VLM backbone & Model & $\textit{TS}_\text{gp}$ $\uparrow$& $\textit{TS}_\text{rp}$ $\uparrow$& $\textit{TS}_\text{ds}$ $\uparrow$& $\textit{TS}_\text{avg}$ $\uparrow$
            \\
			\midrule
			LLaVA v1.5 Baseline & -  & $28.17$ & $19.30$ & $42.06$ & $23.16$ 
            \\
            Fine-tuned LLaVA v1.5  & NexusAD  \cite{mo2024nexusad} & $68.97$ & $57.58$ & $84.31$ & $65.02$ 
            \\
            Fine-tuned LLaVA v1.5 & llmforad \cite{xue2024twostage} & $72.12$ & $58.70$ & $83.41$ & $74.26$ 
            \\
            \midrule
			Janus Pro Baseline & - & $25.88$ & $52.06$ & $30.90$ & $36.28$ 
            \\
			Fine-tuned Janus Pro & AppleVLM  & $\mathbf{73.14}$ & $\mathbf{59.11}$ & $\mathbf{85.11}$ & $\mathbf{75.12}$ 
            \\
		\bottomrule
		\end{tabular}
	\end{center}
\end{table}

\noindent \textbf{VLM Fine-tuned with Corner Cases:}
As described in Sec.~\ref{sec:ps}, the VLM backbone of AppleVLM is pre-trained with a real-world corner-case dataset before the last-stage end-to-end training. Our intuition is that the model can better generalize in long-tail scenarios by incorporating prior knowledge of corner cases. To verify the conjecture, we conduct comparison experiments on CODA-LM~\cite{li2024automated} and DriveLM~\cite{sima2025drivelm} benchmarks.

Regrading CODA-LM, we consider LLaVA v1.5~\cite{liu2024llavanext} and Janus Pro~\cite{chen2025janus}, as well as two models that are fine-tuned on the LLaVA v1.5, NexusAD~\cite{mo2024nexusad}, llmforad~\cite{xue2024twostage} and our proposed AppleVLM that is fine-tuned on Janus Pro. For a comprehensive analysis, we first use the Text-Score metric \textit{TS} to evaluate the standalone VLMs in three tasks of CoT (Tab.~\ref{tab_results_LLM}) among all models, and then compare the closed-loop driving performance of AppleVLM embedded with the vanilla or fine-tuned Janus Pro VLM backbone (Tab.~\ref{tab_results_fine_tuning_applied}). It can be found from Tab.~\ref{tab_results_LLM} that in general, all the models fine-tuned with corner cases (\ie, NexusAD, llmforad, and AppleVLM) significantly outperform the baselines without fine-tuning in text prediction among three CoT tasks. AppleVLM achieves the best \textit{TP} results of $73.14$, $59.11$ and $85.11$ in all three tasks, showing the highest gains ($+47.26$ and $+54.21$) in $\textit{TP}_\text{gp}$ and $\textit{TP}_\text{ds}$ after fine-tuning. This demonstrates that incorporating extreme scenarios in training can enhance the ability of VLMs to understand driving scenes.

On DriveLM benchmark, we consider two VLM baselines (BLIP2 and DriveLM-agent) that provided by~\cite{sima2025drivelm}. As shown in Tab.~\ref{tab_results_Spatial}, AppleVLM significantly outperforms BLIP-2 and DriveLM-agent on SPICE, GPT Score, and Completeness metrics, achieving the best results of $45.45$, $72.04$, and $31.15$. Besides, AppleVLM surpasses the vanilla Janus Pro by $+7.06$, $+9.89$, and $+4.13$, respectively. These results demonstrate that the AppleVLM incorporating the CoT mechanism can effectively improve its spatial-temporal understanding capability.

We further explore the impact of CoT fine-tuning on the final closed-loop driving performance. We provide three settings of two VLM backbones (LLaVA v1.5 and Janus Pro): VLM baselines without fine-tuning, VLMs fine-tuned without CoT (\textit{i.e.,} simply outputing driving suggestions), and VLMs fine-tuned with CoT. As shown in Tab.~\ref{tab_results_fine_tuning_applied}, we see that fine-tuning VLMs is effective in improving end-to-end driving. Whether using LLaVA v1.5 or Janus Pro, CoT-based fine-tuning consistently outperforms simple-task fine-tuning, demonstrating the clear advantage of incorporating the CoT mechanism during the fine-tuning process.


\begin{table}[!t]
	\begin{center}
        \fontsize{8pt}{8pt}\normalfont  
		\caption{The experiments of models conducted on the DriveLM~\cite{sima2025drivelm} benchmark. The $\uparrow$ stands for the higher the better. For each metric, the best result is shown in bold.}
		\label{tab_results_Spatial}  
        \setlength{\tabcolsep}{3pt}
		\begin{tabular}{cccc}
			\toprule 
			VLM Model & SPICE $\uparrow$ & GPT Score $\uparrow$ & Completeness $\uparrow$ \\
			\midrule
			BLIP-2 & 4.336 & 42.97 & 1.064  \\
            DriveLM-agent & 42.56 & 71.39 & 30.04 \\
            \midrule
			LLaVA v1.5 Baseline & 31.85 & 53.51 & 21.43 \\
			Janus Pro Baseline & 38.39 & 62.15 & 27.02 \\
			AppleVLM & \textbf{45.45} & \textbf{72.04} & \textbf{31.15} \\
		\bottomrule
		\end{tabular}
	\end{center}
\end{table}

\begin{table}[!t]
	\begin{center}
        \fontsize{8pt}{8pt}\normalfont  
		\caption{Ablation study on the impact of CoT fine-tuning for the final driving performance. The models are fine-tuned with both the DriveLM~\cite{sima2025drivelm}and CODA-LM~\cite{li2024automated} datasets. The $\uparrow$ stands for the higher the better. For each metric, the best result is shown in bold.}
		\label{tab_results_fine_tuning_applied}  
        \setlength{\tabcolsep}{2pt}
		\resizebox{\linewidth}{!}{\begin{tabular}{cc ccc}
			\toprule 
			Backbone & CoT & $\textit{DS}$ $\uparrow$ & $\textit{RC}$ $\uparrow$ & $\textit{IS}$ $\uparrow$ \\
			\midrule
            
            LLaVA v1.5 Baseline 
            & -  & $44.6 \pm 1.1$ & $57.2 \pm 5.3$ & $0.78 \pm 0.01$ \\
            Fine-tuned LLaVA v1.5 & w/o & $49.9 \pm 1.7$ & $60.9 \pm 2.2$ & $0.82 \pm 0.06$ \\
            Fine-tuned LLaVA v1.5 & w/ & $\mathbf{52.5 \pm 1.5}$ & $\mathbf{63.2 \pm 1.3}$ & $\mathbf{0.83 \pm 0.04}$ \\
            
            \midrule
	        Janus Pro Baseline & -  & $48.7 \pm 2.3$ & $60.1 \pm 3.1$ & $0.81 \pm 0.02$ \\
            
            Fine-tuned Janus Pro & w/o & $52.3 \pm 2.1$ & $61.5 \pm 4.3$ & $0.85 \pm 0.03$ \\
            
            Fine-tuned Janus Pro & w/ & $\mathbf{59.2 \pm 1.9}$ & $\mathbf{67.1 \pm 3.4}$ & $\mathbf{0.89 \pm 0.04}$ \\
		\bottomrule
		\end{tabular}}
	\end{center}
\end{table}

\begin{table}[!t]
	\begin{center}
		\caption{Ablation study on the impact of different datasets applied in CoT fine-tuning for VLMs driving performance. The $\uparrow$ stands for the higher the better. For each metric, the best result is shown in bold.}
		\label{tab_results_two_LLM}  
        \setlength{\tabcolsep}{2pt}
        \begin{tabular}{c cc ccc}
            \toprule 
            Backbone & CODA-LM & DriveLM & $\textit{DS}$ $\uparrow$ & $\textit{RC}$ $\uparrow$ & $\textit{IS}$ $\uparrow$ \\
            			\midrule
            \multirow{4}{*}{LLaVA v1.5}& - & - & $44.6 \pm 1.1$ & $57.2 \pm 5.3$ & $0.78 \pm 0.01$ \\
            & \checkmark & - & $50.4 \pm 1.5$ & $61.5 \pm 1.7$ & $0.82 \pm 0.04$ \\
     	    & -  & \checkmark & $51.4 \pm 1.6$ & $62.7 \pm 1.9$ & $0.82 \pm 0.04$ \\
            & \checkmark & \checkmark & $\mathbf{52.5 \pm 1.5}$ & $\mathbf{63.2 \pm 1.3}$ & $\mathbf{0.83 \pm 0.04}$ \\
                \midrule
            \multirow{4}{*}{Janus Pro}& - & - & $48.7 \pm 2.3$ & $60.1 \pm 4.5$ & $0.81 \pm 0.02$ \\
    	    & \checkmark & - & $55.0 \pm 2.2$ & $63.5 \pm 5.3$ & $0.88 \pm 0.04$ \\
     	    & - & \checkmark & $56.6 \pm 1.7$ & $65.5 \pm 3.1$ & $0.88 \pm 0.04$ \\
            & \checkmark & \checkmark & $\mathbf{59.2 \pm 1.9}$ & $\mathbf{67.1 \pm 3.4}$ & $\mathbf{0.89 \pm 0.04}$ \\
    		\bottomrule
        \end{tabular}
        
	\end{center}
\end{table}
\begin{table*}[!t]
	\begin{center}
		\caption{Ablation study on the impact of freezing or fine-tuning the vision encoder. The models are tested on LangAuto~\cite{shao2024lmdrive} benchmark. The $\uparrow$ stands for the higher the better. The best result is shown in bold.}
		\label{tab_results_freezing_results}  
            \begin{tabular}{c ccc cc}
            \toprule
            Vision Encoder Status & $\textit{DS}$ $\uparrow$ & $\textit{RC}$ $\uparrow$ & $\textit{IS}$ $\uparrow$ & \makecell{Trainable \\ Params (M)} & \makecell{Training Time \\ (h/epoch)} \\
            \midrule
            freezing & $60.5 \pm 2.1$ & $68.7 \pm 3.7$ & $0.88 \pm 0.02$ & $\mathbf{23}$ & $\mathbf{6.5}$ \\
            jointly fine-tuning  & $\mathbf{61.5 \pm 1.8}$ & $\mathbf{70.5 \pm 2.6}$ & $\mathbf{0.88 \pm 0.01}$ & 127 & 9 \\
            \bottomrule
            \end{tabular}
	\end{center}
\end{table*}

\begin{table*}[!t]
\centering
\caption{
Ablation study on the contributions of different modalities to end-to-end autonomous driving performance. The models are evaluated on the LangAuto~\cite{shao2024lmdrive} benchmark. $\uparrow$ indicates higher is better. For each metric, the best result is shown in bold.}
\renewcommand{\arraystretch}{1.2}
\setlength{\tabcolsep}{5pt}
\begin{tabular}{cccc|ccc}
\hline
\multicolumn{4}{c|}{Input Modality} & \multicolumn{3}{c}{Performance} \\
 \cline{1-7}
Image & LiDAR & BEV tokens & Language & $DS$ $\uparrow$ & $RC$ $\uparrow$& $IS$ $\uparrow$\\
\hline
\checkmark & \checkmark & \checkmark & \checkmark & $\mathbf{59.2 \pm 1.9}$ & $\mathbf{67.1 \pm 3.4}$ & $\mathbf{0.89 \pm 0.04}$ \\
\checkmark &  &  \checkmark & \checkmark & $33.6 \pm 1.8$ & $55.1 \pm 1.3$ & $0.61 \pm 0.04$ \\
& \checkmark & \checkmark & \checkmark & $29.8 \pm 1.9$ & $52.2 \pm 2.1$ & $0.57 \pm 0.02$ \\
\checkmark & \checkmark &  & \checkmark & $41.3 \pm 1.9$ & $59.9 \pm 1.7$ & $0.69 \pm 0.01$ \\
\checkmark & \checkmark & \checkmark &  & $31.9 \pm 2.0$ & $47.2\pm 1.9$ & $0.59 \pm 0.01$ \\
& & & \checkmark & $2.6 \pm 0.1$ & $3.2\pm 0.1$ & $0.80 \pm 0.01$ \\
\hline
\end{tabular}
\label{r1_c3}
\end{table*}

\noindent \textbf{CoT fine-tuning datasets:} In Tab.~\ref{tab_results_two_LLM}, we analyze the impact of different datasets used for CoT fine-tuning of the VLM backbones on the final driving performance. We observe that either CODA-LM and DriveLM CoT fine-tuning for both LLaVA v1.5 and Janus Pro VLMs has a positive contribution to improve the model’s driving performance. Compared to CODA-LM, DriveLM yields a slight advantage in performance gain. The best results are achieved when using both datasets.

\noindent \textbf{Freezing Vision Encoder:} As shown in Tab.~\ref{tab_results_freezing_results}, we study the impact of freezing or fine-tuning vision encoder in the training process of driving task. We observe that fine-tuning the visual encoder yields only marginal gains ($+1.8$ in \textit{RC} and $+1.0$ in \textit{DS}), yet prolongs training time and increases the number of trainable parameters by $5$ times per epoch. To strike a balance between performance and efficiency, in this work, we freeze the pre-trained vision encoder when training the driving task.

\noindent \textbf{Input Modality:} As shown in Tab.~\ref{r1_c3}, we analyze the impact of different modalities on the model driving performance, evaluated on the LangAuto benchmark. We see that incorporating all the vision, language, and planning modules contributes to the best driving performance of AppleVLM. Removing the LiDAR or RGB image modality significantly degrades \textit{DS}, \textit{RC}, and \textit{IS} results. With only the language input, the model is nearly incapable of performing the driving task. This aligns with the fundamental role of human driving, where vision perception is essential. In addition, both the BEV tokens (planning) and the language modalities play an important role in ensuring strong driving performance.

\begin{figure*}[!t]
    \centering
    \includegraphics[width=0.8\linewidth]{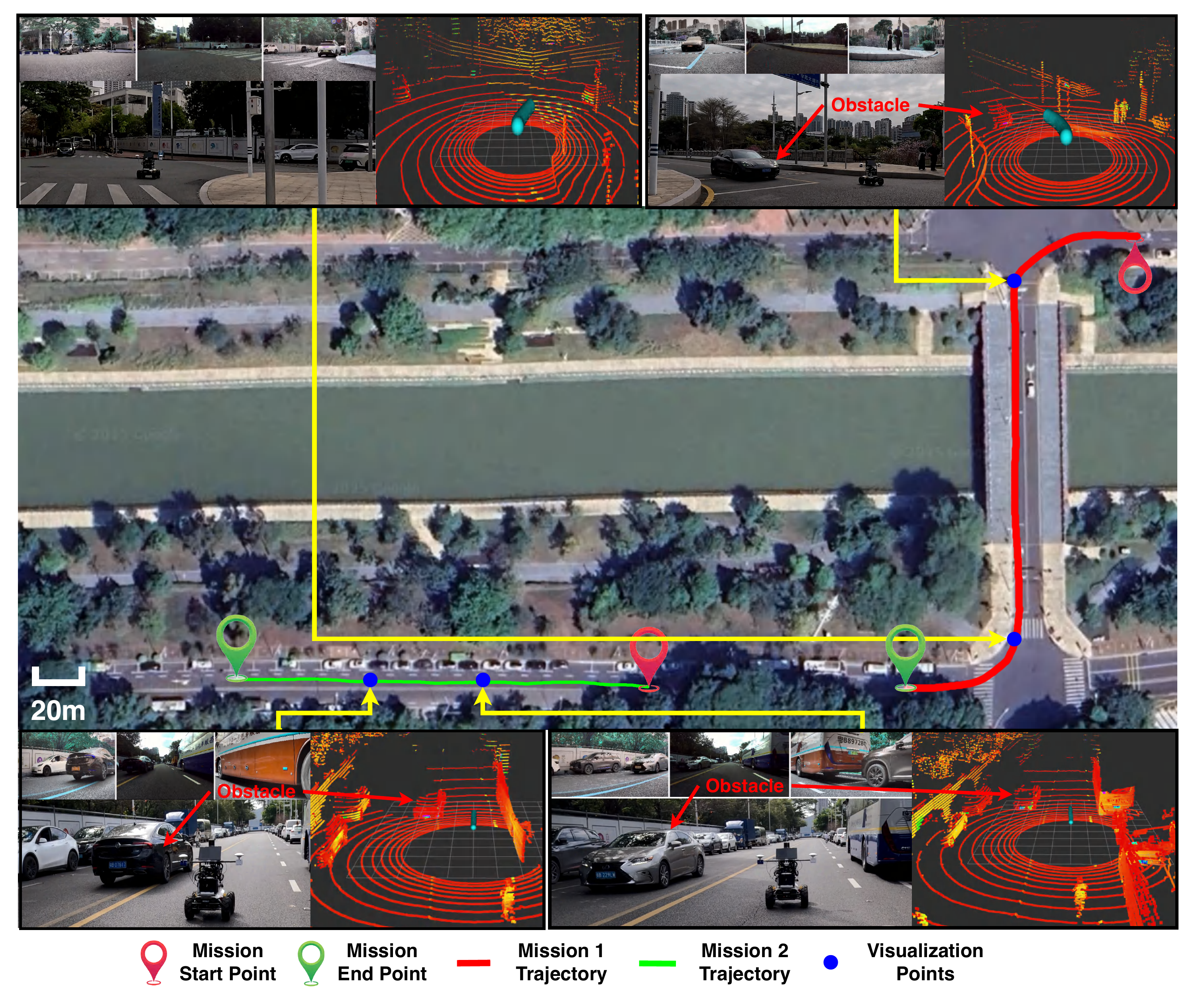}
    \caption{The closed-loop driving in the real-world outdoor environment. The green line indicates the driving trajectory of \textit{Going Straight} task, while the red one represents the driving trajectory of \textit{Turning} task. Several waypoints (blue dots) are zoomed in to show the real driving environment (with dynamic pedestrians and vehicles around the ego vehicle).}
    \label{fig_closed_loop}
\end{figure*}

\subsection{Real-world Deployment} \label{sec_rwd}
To validate the generalization capability of AppleVLM in the real world, we instantiate our model on an AGV platform without any fine-tuning by real-world driving data. Moreover, the sensor configurations are distinct from the ones used on the training data collected from the CARLA simulator. For safety concerns, we perform two handy closed-loop end-to-end driving tasks with dynamic obstacles such as random cyclists, pedestrians and vehicles around: 1) a $185$ meters straight-line driving; 2) a $230$-meter route includes both left and right turning. The end-to-end inference runs at $120$ ms per frame ($\sim$$8$ FPS) on a single NVIDIA AGX Orin. The GPU utilization remains below $60$\% and the memory consumption is around $10$ GB, which demonstrates the model’s efficiency on embedded hardware.

As demonstrated in Fig.~\ref{fig_closed_loop}, AppleVLM can complete two closed-loop end-to-end driving tasks without human takeover. Our model maintains driving without being disturbed by the surrounding obstacles. We provide a video in the supplementary materials to demonstrate the driving of AppleVLM in the real world.
Note that there exists a significant domain gap between the CARLA simulation and the real-world environments, including object appearance, data distribution, sensor configuration, {\etc}, therefore, for learning-based models, sim-to-real has always been a critical yet unresolved challenge. Despite this, AppleVLM demonstrates promising generalization capability. Given that our method is simple and flexible, it would be interesting to explore it further in more complex driving tasks and scenarios with advanced domain adaptation techniques.

\section{Conclusion}
In this work, we propose a novel end-to-end model AppleVLM, that incorporates three modalities: vision, language, and planning for autonomous driving. 
We leverage the deformable transformer mechanism to enhance the vision encoder, enabling AppleVLM to be robust to variations in sensor configurations between training and real-world deployment. By fusing RGB images and point-cloud data in both spatial and temporal aspects, AppleVLM achieves higher driving and infraction scores on two CARLA driving benchmarks and shows a safer driving performance with fewer collisions.
Compared to the traditional VLMs, we add the new planning modality to encode environmental spatial information provided by explicit BEVs that reduce language biases in navigation instruction.
Furthermore, by fine-tuning the VLM backbone using corner cases through a CoT training mechanism, AppleVLM can effectively improve the generalization ability in out-of-distribution scenarios. 
Finally, real-world experiments validate the feasibility of directly deploying AppleVLM, trained in simulation, onto real vehicles without requiring domain adaptation.

\bibliographystyle{IEEEtran}
\bibliography{IEEEabrv,TITS-final}
\end{document}